\documentclass[10pt,twocolumn,letterpaper]{article}

\usepackage{iccv}
\usepackage{times}
\usepackage{epsfig}
\usepackage{graphicx}
\usepackage{amsmath}
\usepackage{amssymb}

\usepackage{multirow}
\usepackage{array}
\usepackage{bm}

\newcommand{\bsls}{\textbackslash}

\usepackage[pagebackref=true,breaklinks=true,letterpaper=true,colorlinks,bookmarks=false]{hyperref}

\iccvfinalcopy

\def\iccvPaperID{1634} 

\ificcvfinal\fi
\begin{document}

\title{CrowdCam: Dynamic Region Segmentation}

\author{Nir Zarrabi\\
Tel-Aviv University, Israel\\
{\tt\small nirz1@mail.tau.ac.il}
\and
Shai Avidan\\
Tel-Aviv University, Israel\\
{\tt\small avidan@eng.tau.ac.il}
\and
Yael Moses\\
The Interdisciplinary Center, Israel\\
{\tt\small yael@idc.ac.il}
}

\maketitle

\begin{abstract}
We consider the problem of segmenting dynamic regions in CrowdCam images, where a dynamic region is the projection of a moving 3D object on the image plane. Quite often, these regions are the most interesting parts of an image. CrowdCam images is a set of images of the same dynamic event, captured by a group of non-collaborating users. Almost every event of interest today is captured this way. This new type of images raises the need to develop new algorithms tailored specifically for it.

We propose a comprehensive solution to the problem. Our solution combines cues that are based on geometry, appearance and proximity. First, geometric reasoning is used to produce rough score maps that determine, for every pixel, how likely it is to be the projection of a static or dynamic scene point. These maps are noisy because CrowdCam images are usually few and far apart both in space and in time. Then, we use similarity in appearance space and proximity in the image plane to encourage neighboring pixels to be labeled similarly as either static or dynamic. 

We collected a new, and challenging, data set to evaluate our algorithm. Results show that the success score of our algorithm is nearly double that of the current state of the art approach.

\end{abstract}
\section{Introduction}
\begin{figure}[t]
\begin{center}
\def\arraystretch{1.5}
{\setlength{\tabcolsep}{3pt}
\begin{tabular}{lll}

\includegraphics[width = .305\linewidth]{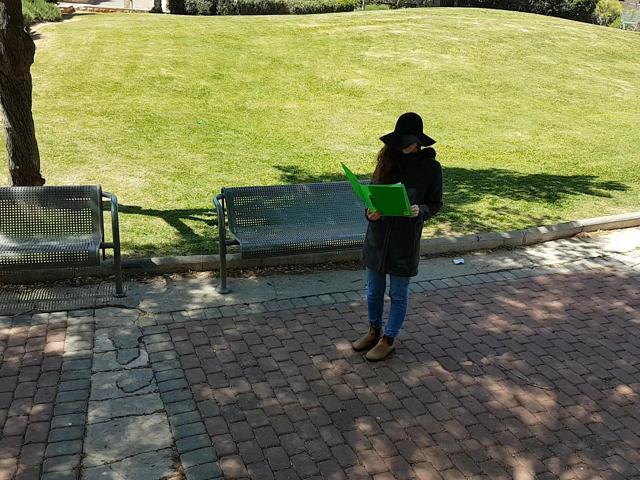} & 
\includegraphics[width = .305\linewidth]{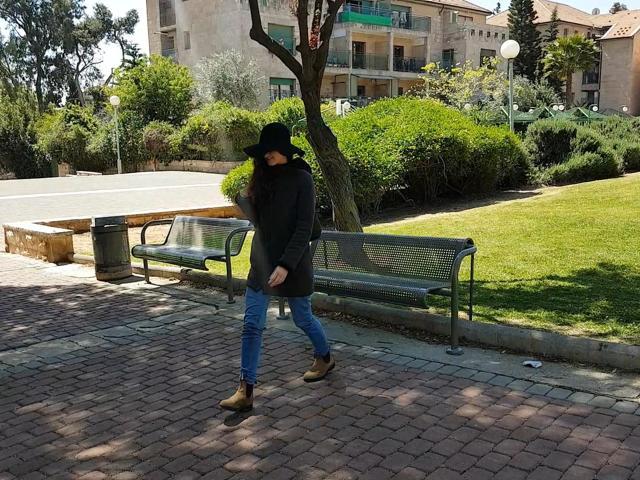} & 
\includegraphics[width = .305\linewidth]{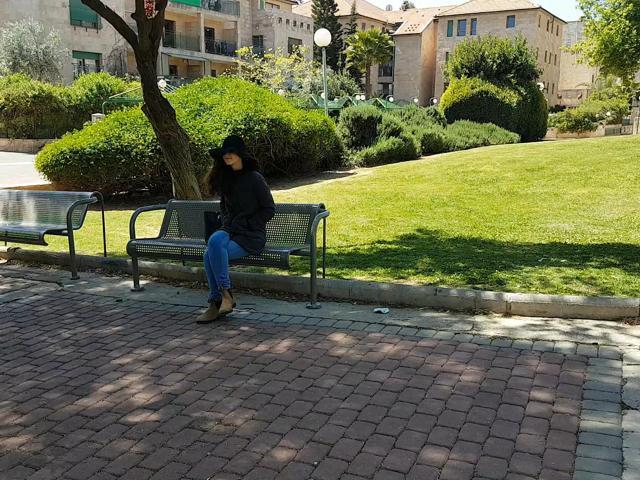} \\
\includegraphics[width = .305\linewidth]{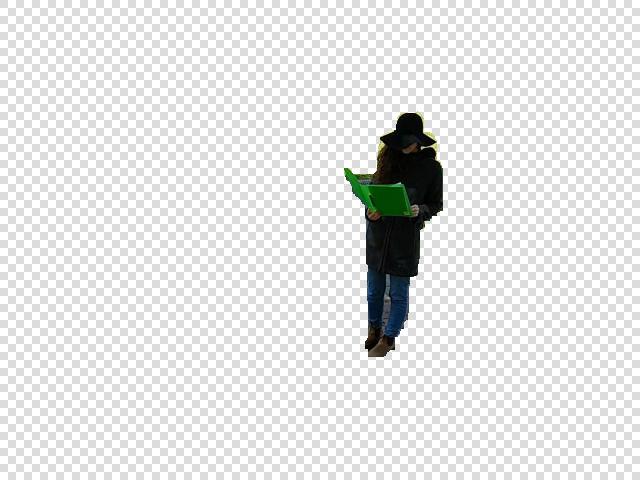} & 
\includegraphics[width = .305\linewidth]{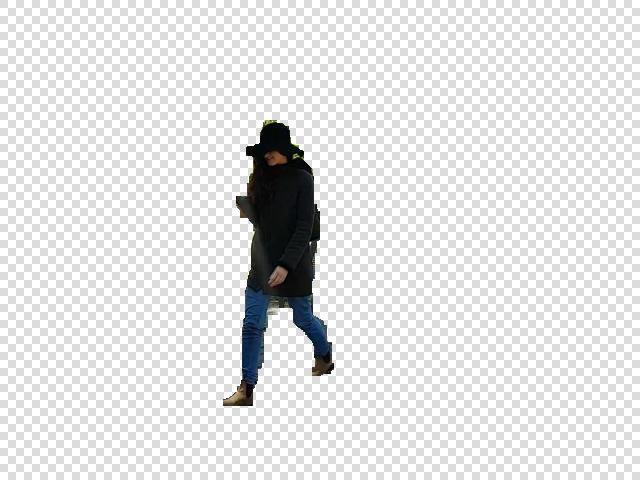} & 
\includegraphics[width = .305\linewidth]{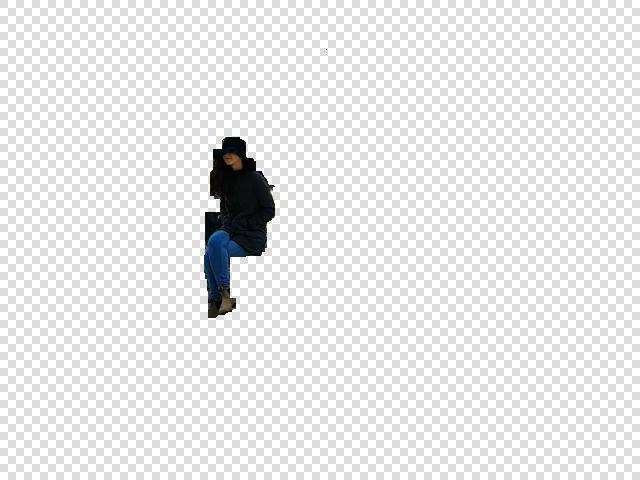}\\

\end{tabular}}
\caption{CrowdCam images segmentation: dynamic regions segmentation of images from the same scene, taken from different viewpoints and captured by multiple cameras. The bottom row presents the segmentation obtained with our method.}
\label{fig:crowdCamImagesIntro}
\end{center}
\end{figure}

CrowdCam images are a collection of still images that capture a dynamic event. Such data is captured nowadays at almost every event of interest. The images are captured by different people with little or no coordination or synchronization (see examples in Fig.~\ref{fig:crowdCamImagesIntro}). We wish to detect and segment the dynamic regions in each of the images, independent of their class and shape. Detecting dynamic regions helps determine where the action is and highlight changes in a scene. The main challenge in detecting dynamic regions is to determine whether the image regions vary due to motion or due to CrowdCam characteristics (i.e., occlusions, illumination variance, viewpoint change, etc.). 

The problem of detecting dynamic regions in CrowdCam images was first addressed by Dafni~\etal~\cite{dafni2017detecting}. Their method produces a {\em dynamic score} map for each image. The  dynamic score of each pixel reflects whether the pixel is a projection of a dynamic or static point in the scene. The score is computed independently for each pixel in each of the images, using only geometric reasoning (as described in Sec.~\ref{InitialMaps}). Given the dynamic score map, a dynamic region segmentation can then be obtained by a na\"{i}ve algorithm that  applies a threshold to each of the score maps.

We propose an overall solution to the problem of dynamic region segmentation in multiple images. Our solution combines the rough geometric cues provided by~\cite{dafni2017detecting} with new cues that are based on similarity in appearance of patches within and across images, as well as proximity between pixels in the image plane. The appearance cues help establish long range relationships between pixels and the proximity cues capture short range relationships.

All the cues (geometry, proximity and appearance) are tied together using a Markov Random Field (MRF).
The unary term is based on a mixture of geometric and appearance cues, and the binary term, that relies on proximity, helps enforce the spatial smoothness assumption. This way, the probability distribution function of the unary term captures long-range pixel interactions and the MRF solver mitigates local errors and adjusts pixel labeling to comply with strong edges in the image. The result of the MRF is an updated dynamic score map for each image. To obtain the final segmentation result, a  threshold can be applied to the computed maps. We refer to our algorithm as CrowdCam Dynamic Region Segmentation, or CDRS.

The dynamic region segmentation task we address differs from semantic segmentation, co-segmentation,  background subtraction, and motion segmentation in video sequences. Semantic segmentation, e.g. segmentation of a car or a person, does not indicate whether the object is dynamic or static. For example, a car may move or park. Motion segmentation in video relies on the proximity in space and time of the video frames, and co-segmentation methods do not assume that objects are dynamic. 
In our scenario, we wish to segment moving objects which might differ between the different images (e.g., a moving car in one view and a dog in another).
Background subtraction methods are also not applicable since CrowdCam images cannot be aligned by homography, due to the wide baseline and non-planar scene, thus the background cannot be learned.  

We evaluate our method on a new and challenging dataset collected by us as well as on the dataset used by~\cite{dafni2017detecting, Kanojia2018}. Experiments show that our method outperforms existing state-of-the-art techniques by one hundred percents.
The main contributions of our paper are: 
(i) Using patches with similar appearance in a set of images of the same scene for  improving dynamic region segmentation;
(ii) Effectively combining the three cues: geometry, proximity and appearance. While each of this components were investigated before, they were never put to use in such a way for this type of problem; 
(iii) New challenging image sets\footnote {Our dataset will soon be publicly available} for evaluating dynamic region segmentation, which can be used for further research of CrowdCam data.

\begin{figure*}[t]
\begin{center}

\includegraphics[width = 1\linewidth]{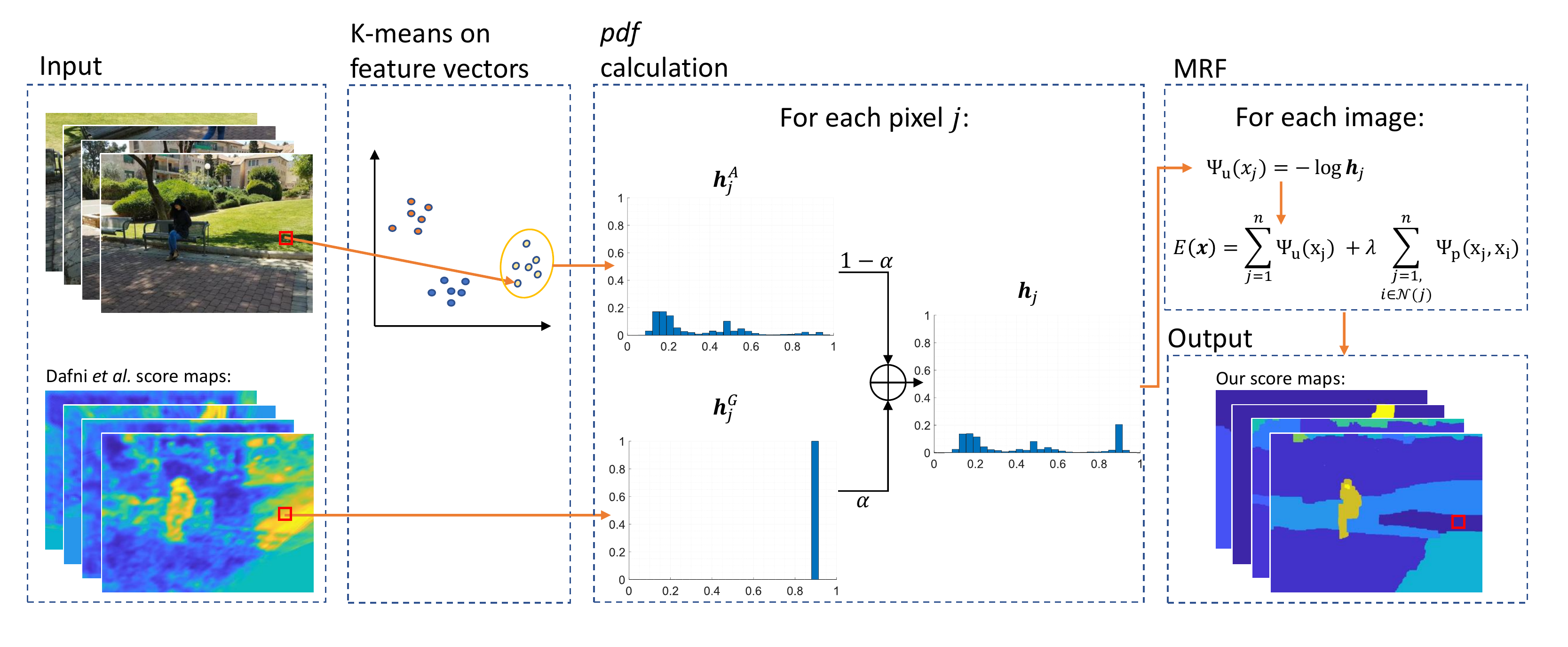} 

\caption{Method outline: the input is a set of CrowdCam images on which we calculate initial score maps using~\cite{dafni2017detecting}. As a second step, we cluster pixel feature vectors using K-means. Next, for each pixel $j$, we calculate a score \textit{pdf}, $h_j$, using a mixture of (i) its initial score value (${\bf h}_j^G$); (ii) the initial score values of all pixels in its cluster (${\bf h}_j^A$). As the last step, we solve an MRF minimization problem to set a final score value for each pixel, using its previously calculated \textit{pdf}.}
\label{fig:MethodDiagram}
\end{center}
\end{figure*}

\section{Background}
Detecting moving regions in CrowdCam images was addressed only in~\cite{dafni2017detecting,Kanojia2018}. Both methods output dynamic score maps, using the observation that matching projections of static regions in the scene must be consistent with epipolar geometry. Our goal is to improve these score maps using proximity as well as appearance cues using patches with a similar appearance in the image itself and in other images of the scene.
 
Co-segmentation of objects, using a set of images, was first introduced by Rother~{\em et al.}~\cite{Rother:2006:Cosegmentation}. Their method, however, does not use motion as a cue. All existing co-segmentation methods use single image cues for segmentation (e.g., color or texture), but they differ in the assumptions they make about other images in the set: the background differs between images (e.g.,~\cite{Rother:2006:Cosegmentation}); the foreground objects are salient in all images (e.g.,~\cite{Rubinstein2013}); or that segmented regions are consistent (e.g.,~\cite{Batra:2010:iCoSeg}). 
None of the above-mentioned methods directly address \textit{moving} region segmentation, as we do in our approach. Nor do we assume that the number of moving regions is known and that the moving objects are necessarily the salient objects in the scene. Our method not only segment the object, but also detect the relevant moving objects. Moreover, we wish to segment moving objects, which are not necessarily present in all images. Note that in CrowdCam images 
the moving objects have the same background. 

Using a set of images can also be used for extracting depth information, which can then be used 
as an additional source of information for object segmentation (e.g.~\cite{jeong2018object, ma2017multi, silberman2012indoor}). Such methods are prone to errors in depth estimation that, in turn, might affect segmentation results. As demonstrated and discussed in~\cite{dafni2017detecting}, existing 3D reconstruction methods fail on CrowdCam images because the images are few and far apart.

Semi-supervised methods use scribbles of foreground or background, provided by the user, for
single image segmentation 
(e.g.~\cite{boykov2001interactive,rother2004grabcut, yu2017loosecut}) or 
    co-segmentation (e.g.~\cite{Batra:2010:iCoSeg,collins2012random, Rother:2006:Cosegmentation}).
An alternative to user scribbles is to use an exact segmentation in one image and propagate it to the rest of the CrowdCam images~\cite{averbuch2018co}. The geometric scores we use can also be regarded as an external input for a segmentation or co-segmentation method. However, the user scribbles are sparse and accurate, whereas our score maps are dense and very noisy.

We use MRF as a basic optimization solver for our task.  MRF was used in many single image segmentation methods  (e.g.,~\cite{rother2004grabcut}) and in  co-segmentation  methods (e.g.,~\cite{Batra:2010:iCoSeg}). 
These methods differ mainly in their definitions of the unary and the binary terms.  
We define a new unary term, that incorporates scores of patches with a similar appearance from all images.
Moreover, our method uses MRF to compute a dynamic score map rather than the final segmentation. 

\section{Method} 
\label{Method}

The input to our method is a set ${\cal I}$ of still CrowdCam images of a dynamic scene,  taken by a set of uncalibrated and unsynchronized cameras from a wide baseline setup. We use a modification of the method of Dafni~\etal~\cite{dafni2017detecting} to compute the  initial score map of each image that reflects whether a pixel is a projection of a static or a dynamic~3D point. Equipped with this initial guess, we propose an MRF solution. Our solution   integrates local information and information across multiple images to obtain better score maps, resulting in a far superior result. The outline of the method is depicted in Fig.~\ref{fig:MethodDiagram}.

\subsection{Initial Map Calculation} 
\label{InitialMaps}
We give a short description of the method of~\cite{dafni2017detecting} for completeness.
They used the observation that corresponding static pixels from multiple images must satisfy the epipolar geometry constraints. This observation is used to compute a per-pixel score that encodes whether the pixel is static. 
 The epipolar geometry (i.e., the fundamental matrix) is computed for a reference image $I\in {\cal I}$, and each image $I'\in {\cal I}$ using computed matching features and the RANSAC method. An initial  static score of a pixel is calculated based on whether there exists at least one potential match along the epipolar lines in $I'\in {\cal I}$. Note that the match is not necessarily the correct one.
   Pixels might not have a match since they are a projection of a moving 3D point, but also due to occlusions, variations in camera parameters, out of the field of view,  etc. Hence,  multiple score maps are calculated, with respect to the other images in ${\cal I}$, and the dynamic score map is obtained by their combination. For more details, see~\cite{dafni2017detecting}.   

The output of this method is an initial score map ${\bf s_r} = \{s_1,\cdots,s_n\}$  for each image $r$, where $n$ is the number of pixels in an image.
The score map   is  computed using the geometric consideration; hence, we regard it as a {\em geometric score}.
  The score value $s_j \in [0, 1]$ encodes whether pixel $j$ is a projection of a static (i.e. $s_j = 0$) or dynamic ($s_j= 1$) $3D$ scene point. 
   

Our implementation differs from that of~\cite{dafni2017detecting} in several important aspects. First, we use DeepMatching~\cite{weinzaepfel:hal-00873592} to compute sparse matches between images. Second, we used deep features to determine patch similarity, in a way similar to~\cite{kat2018matching} and~\cite{talmi2017template}, instead of the HoG features used in~\cite{dafni2017detecting}. The initial feature map for each image was generated by concatenating the output of layers \textit{conv1{\_}2} and \textit{conv3{\_}4} from a pre-trained VGG network~\cite{simonyan2014very}. As the output size of the layers (width and height) differs from that of the input layers, we used bilinear interpolation for resizing. This already improves the original results of~\cite{dafni2017detecting} (see Sec.~\ref{Experiments}).

\begin{figure*}[t]
\begin{center}
\def\arraystretch{1.5}
{\setlength{\tabcolsep}{3pt}

\begin{tabular}{cccccc}

\includegraphics[width = .187\linewidth]{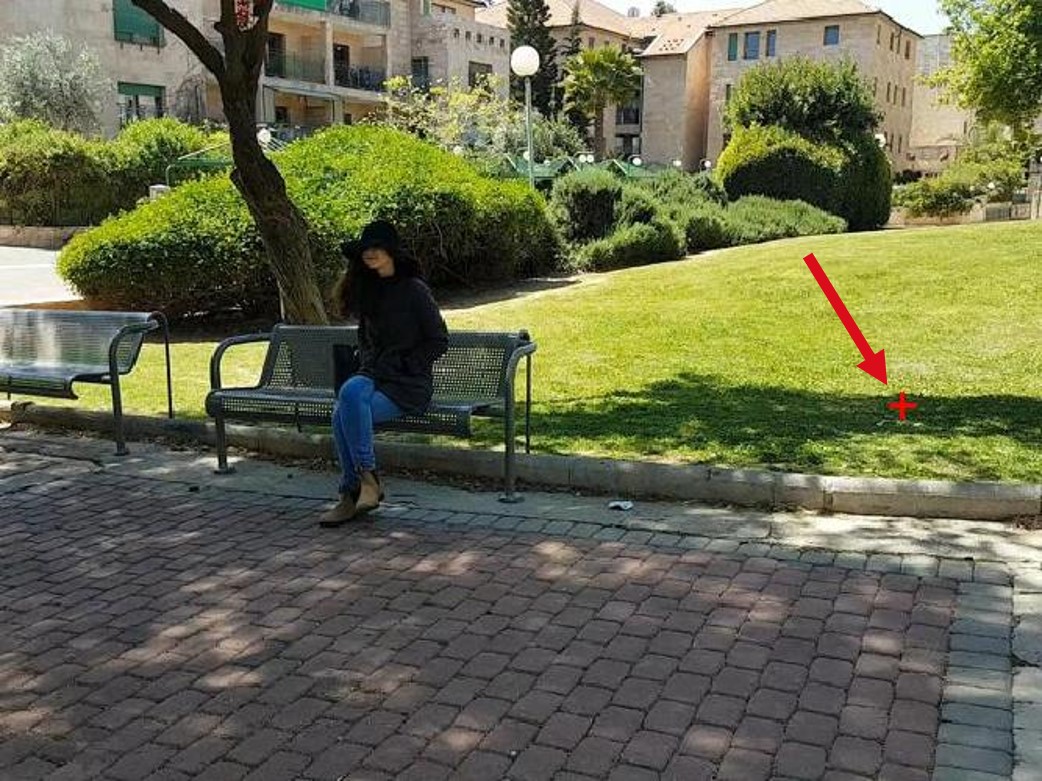}& 
\includegraphics[width = .187\linewidth]{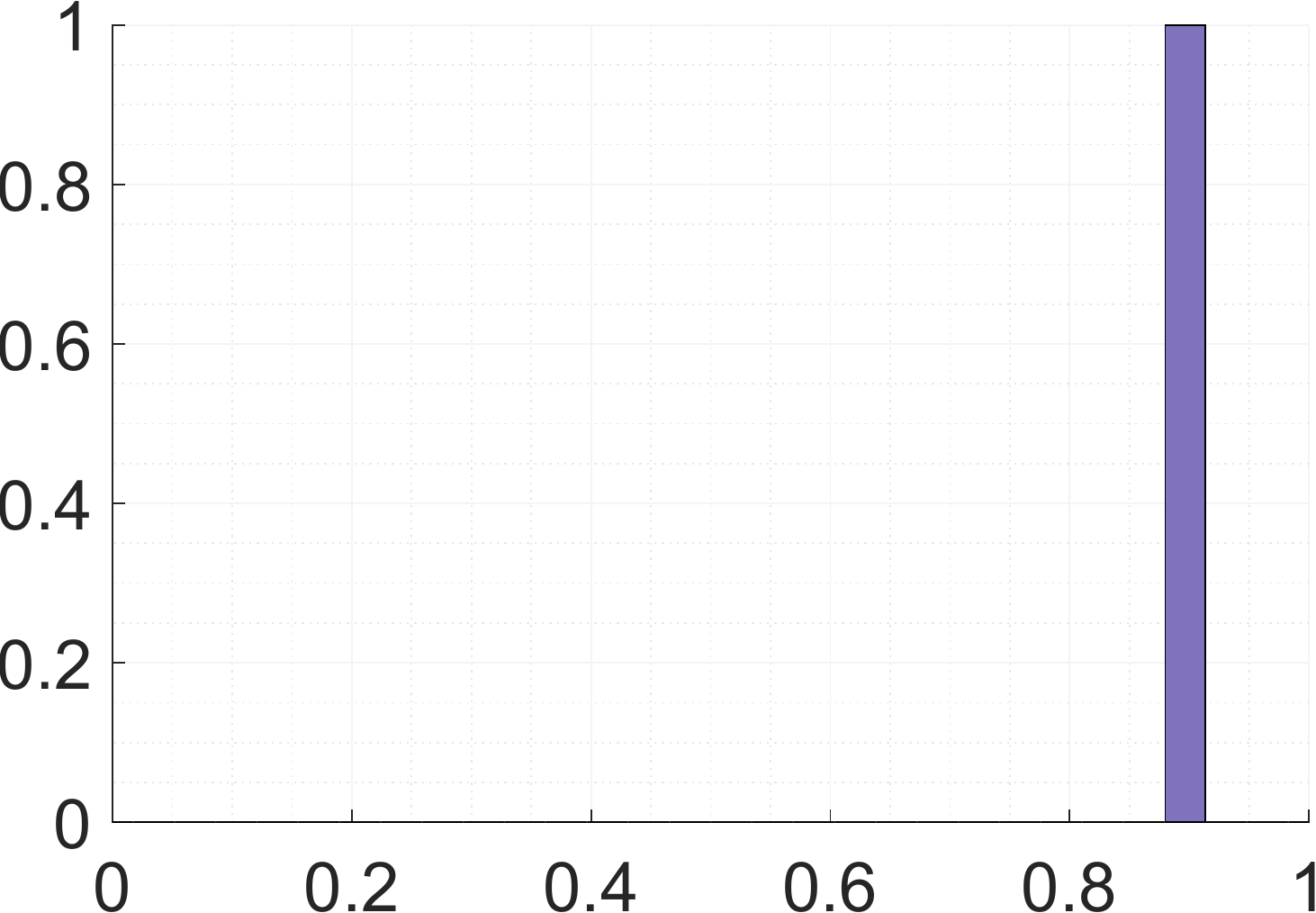}& 
\multirow{-2}{*}{\includegraphics[width = .19\linewidth]{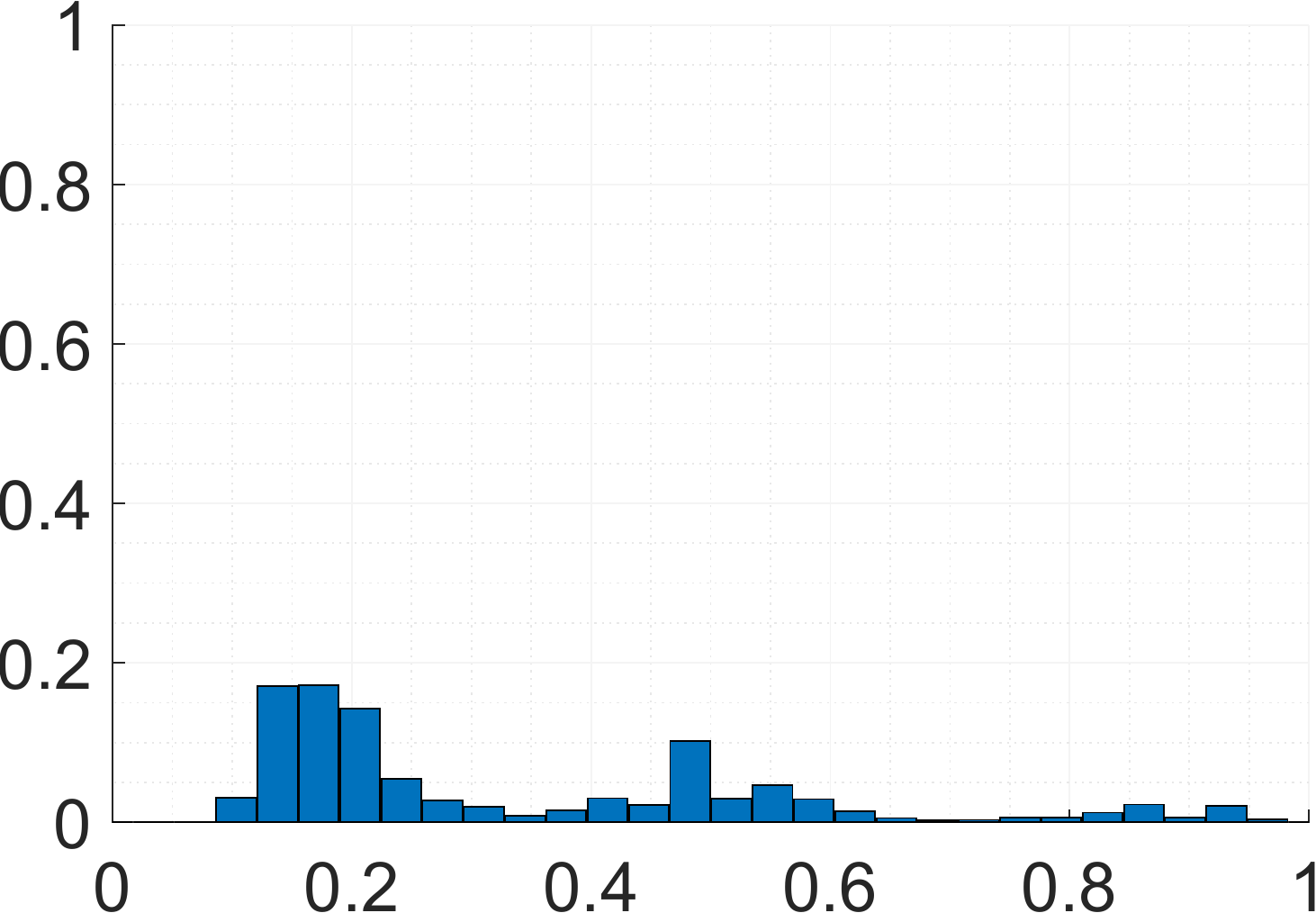}}& 
\includegraphics[width = .187\linewidth]{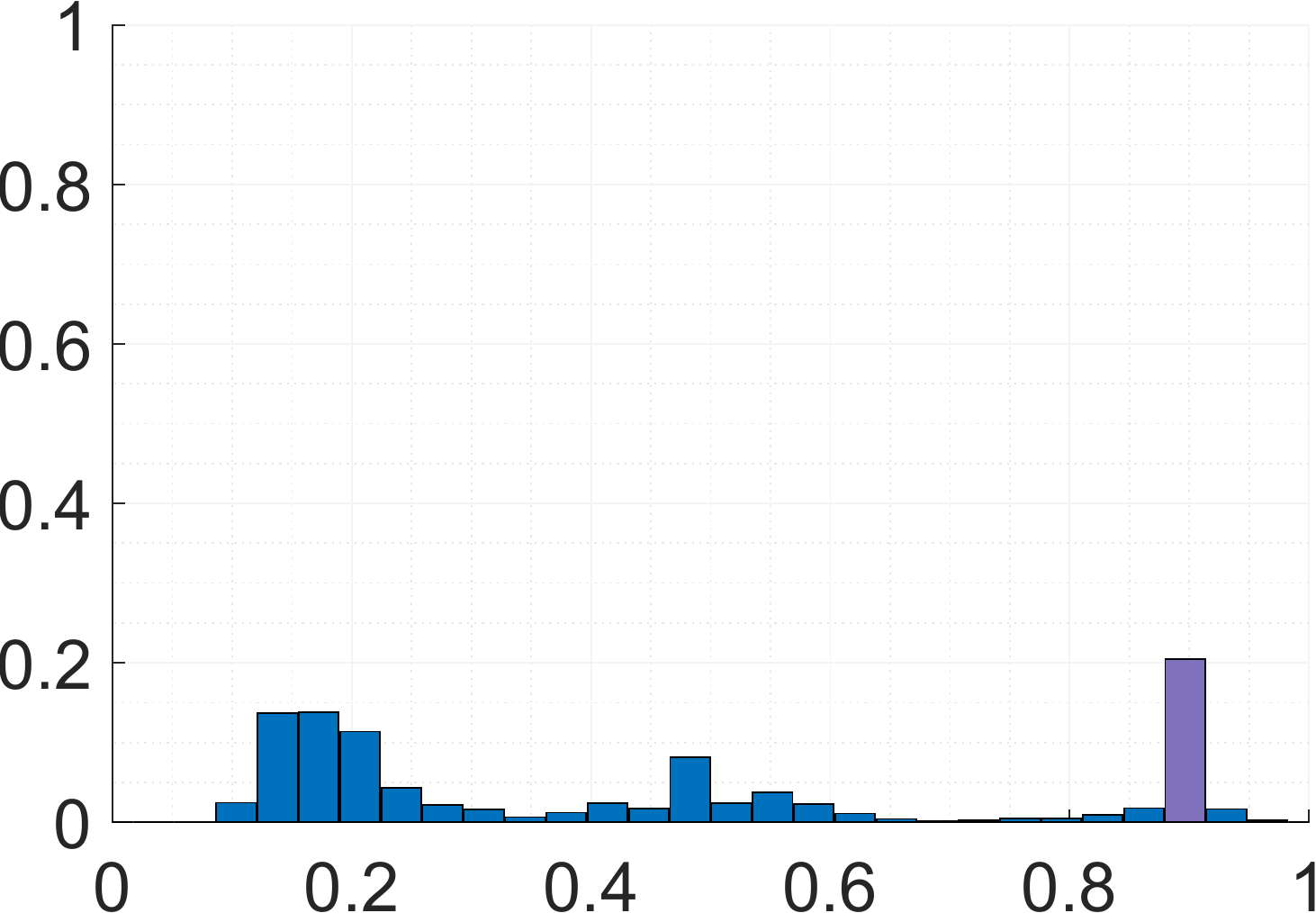} &
\includegraphics[width = .187\linewidth]{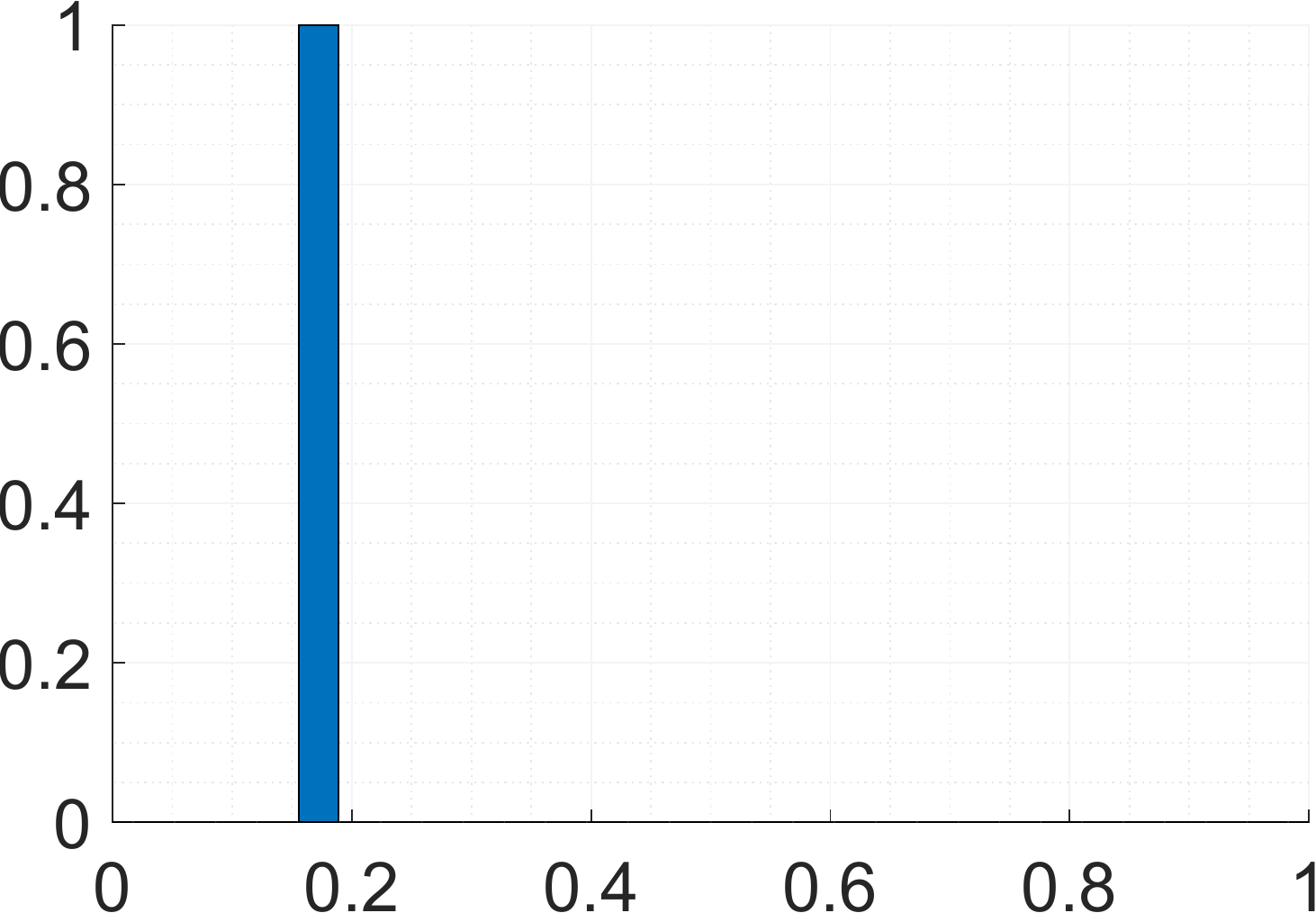}\\

\includegraphics[width = .187\linewidth]{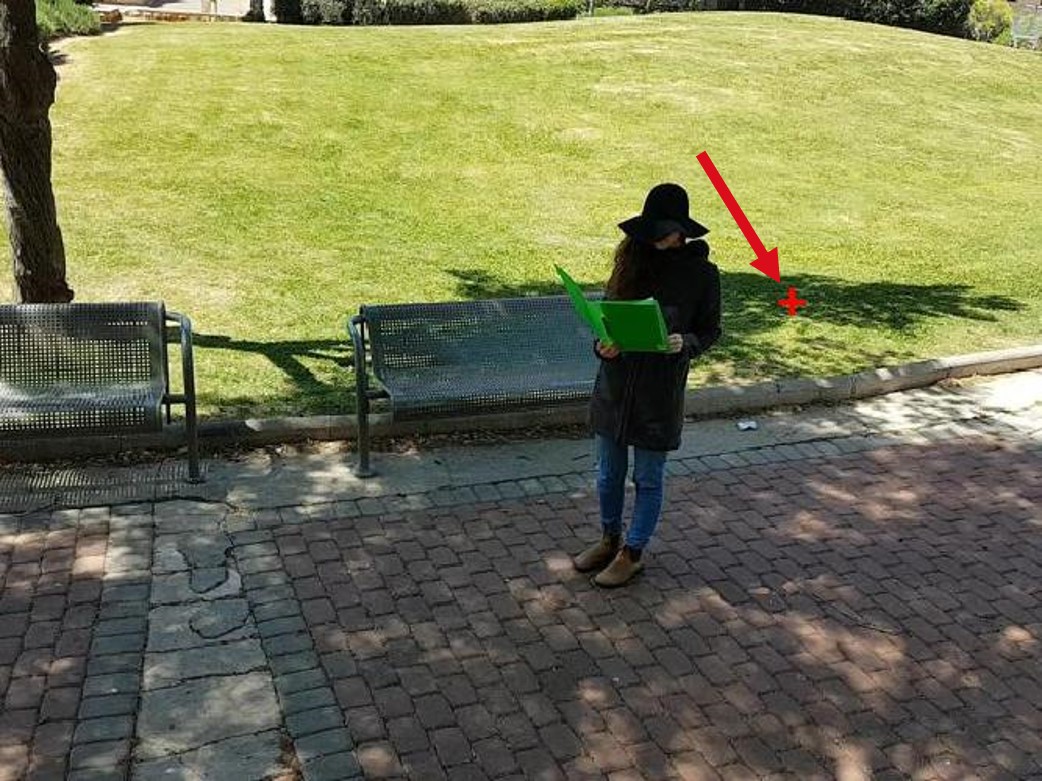}& 
\includegraphics[width = .187\linewidth]{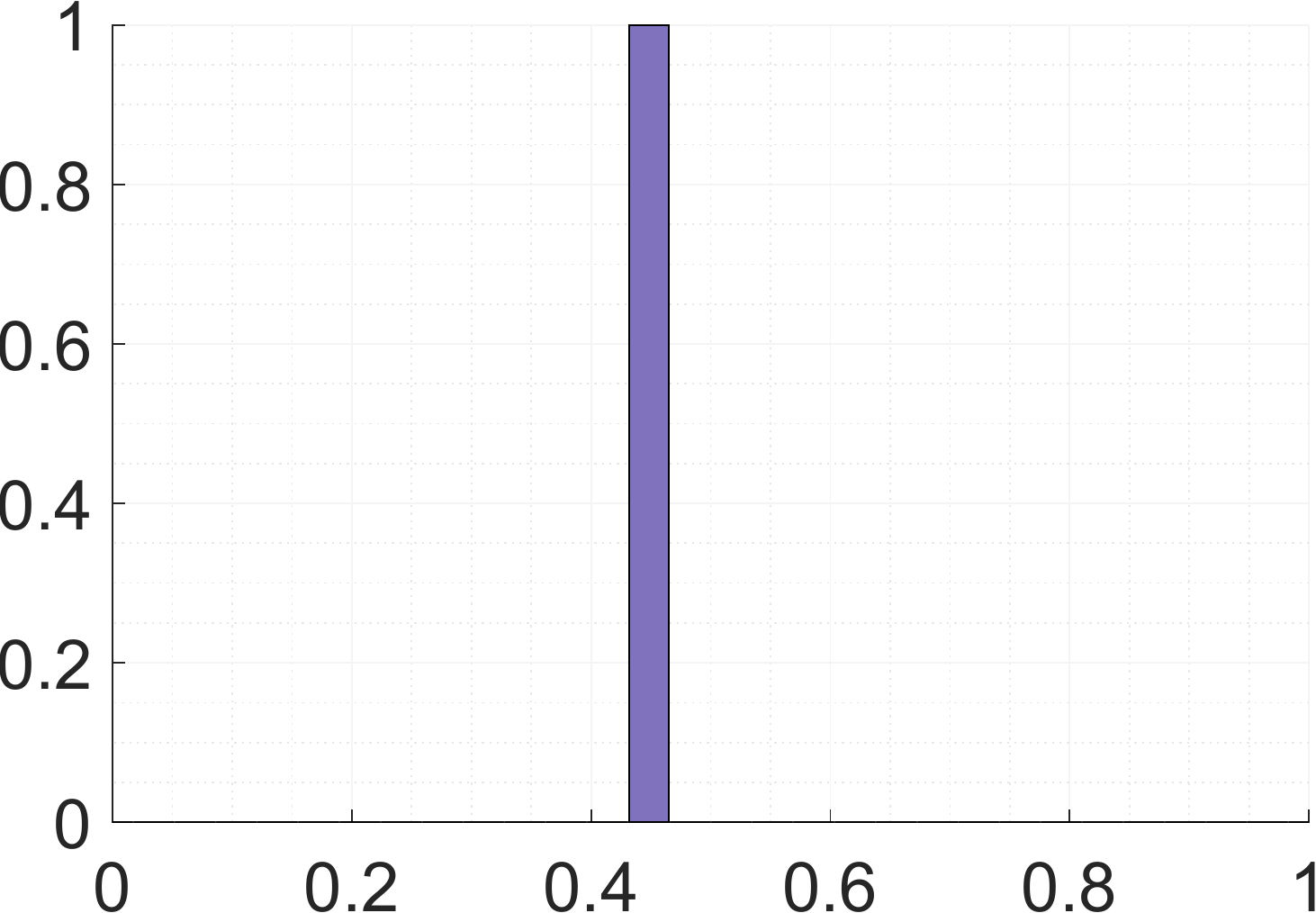}& & 
\includegraphics[width = .187\linewidth]{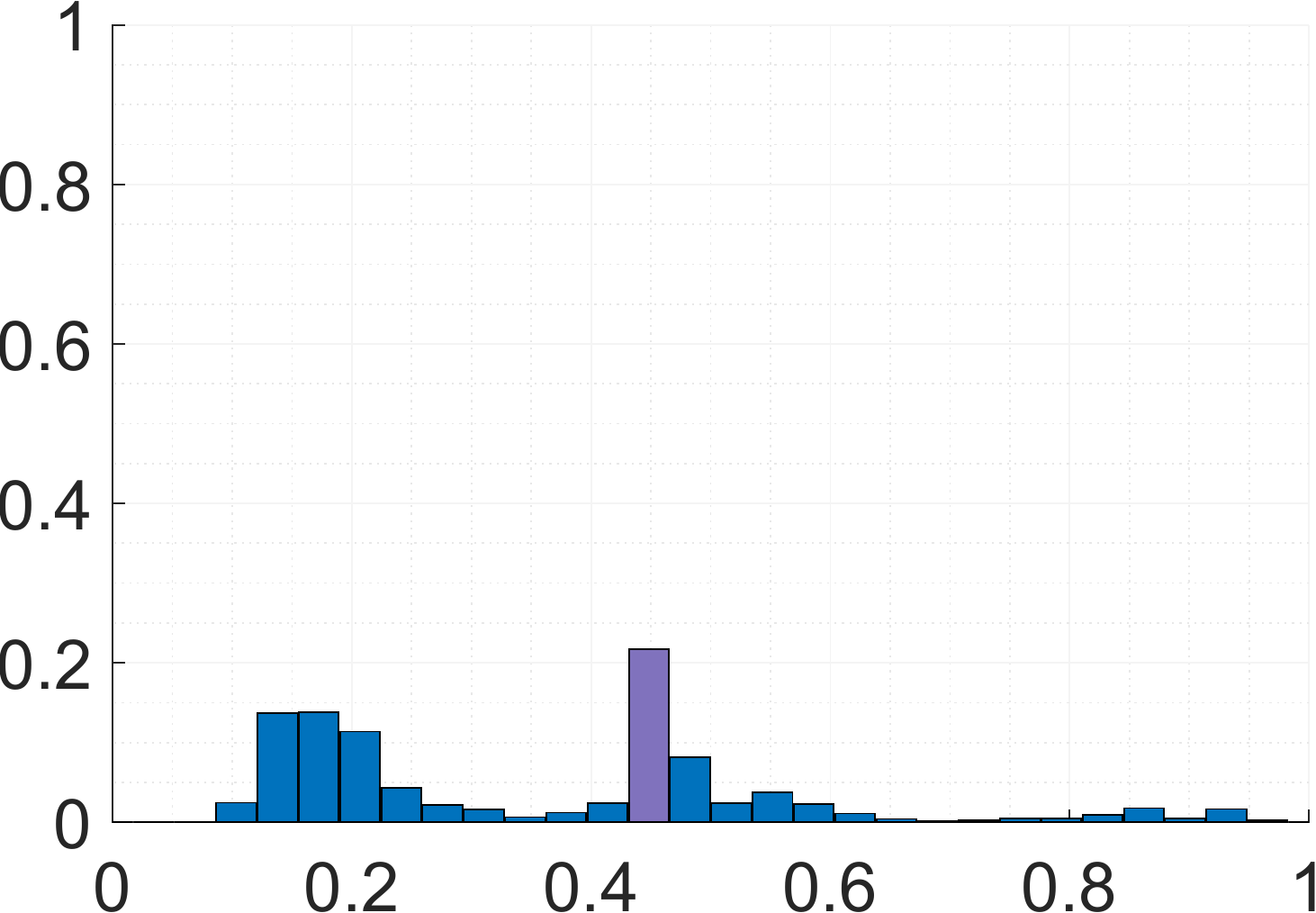} &
\includegraphics[width = .187\linewidth]{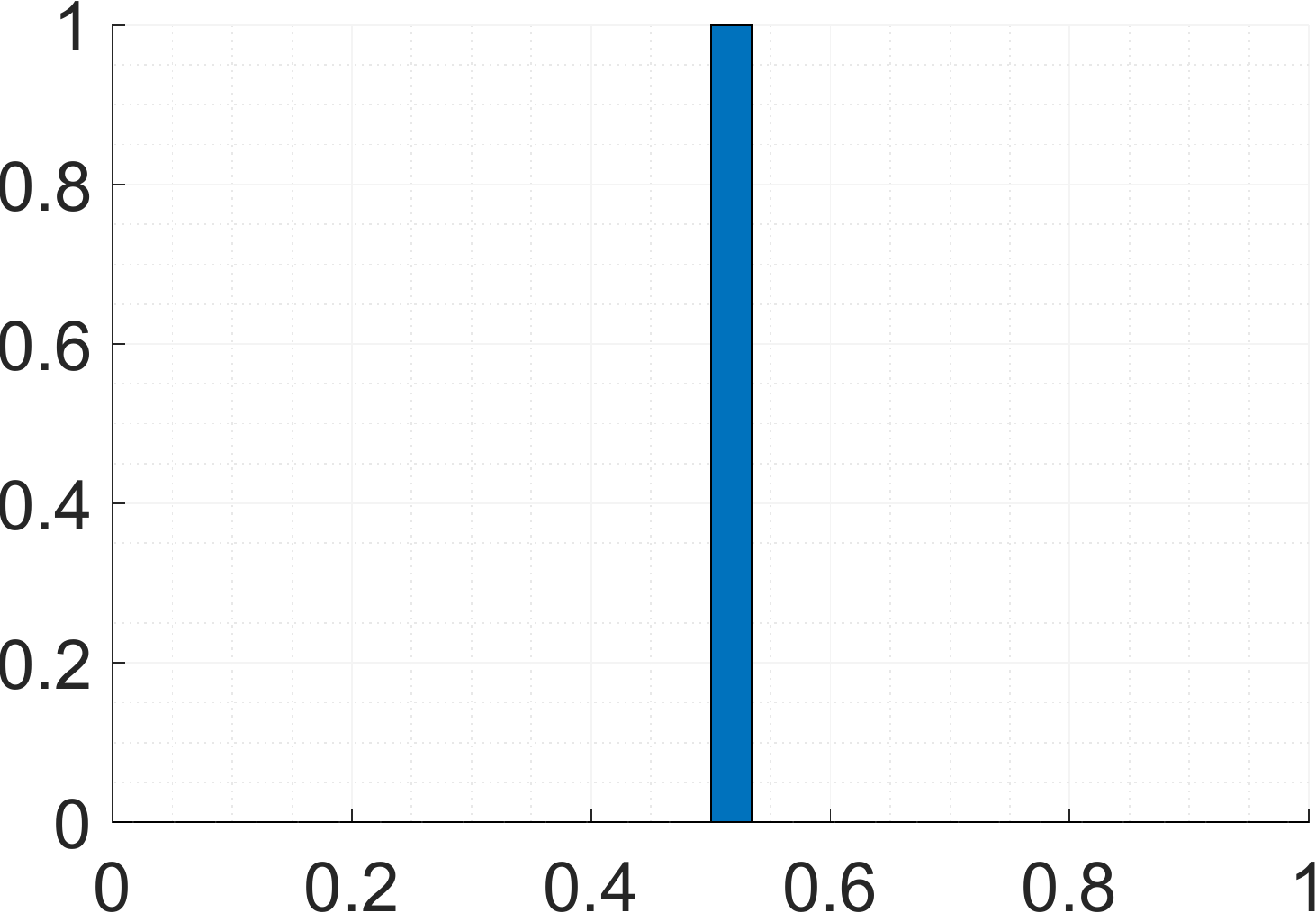}\\
(a) pixels $\{i, j\}$  & (b) ${\bf h}^{G}_{i}, {\bf h}^{G}_{j}$ & (c) ${\bf h}^{A}_{i}, {\bf h}^{A}_{j}$ & (d) ${\bf h}_{i}, {\bf h}_{j}$ & (e) output scores \\

\end{tabular}}
\caption{(a) Two marked pixels $\{i, j\}$ that belong to the same cluster, $\{\textbf{f}_i,\textbf{f}_j\} \in k$. Both pixels correspond to 3D static scene points. Thus, they should have low dynamic scores. (b) Initial geometric score for the marked pixels, i.e., the input to our algorithm. (c) Appearance's  distribution as calculated using Eq.~\ref{eq:pCenter}. (d) Mixture probability distributions. (e) Final pixel scores, i.e. the output of our algorithm after MRF energy minimization.}
\label{fig:figPosteriorProb}
\end{center}
\end{figure*}

\subsection{MRF Solution}

Given a set of maps $\{\bf s_r\}$ as described above, we use a Markov Random Field (MRF) approach to compute a new score map for each image.
A finite number of labels represent the score values (the initial scores are real numbers). Thus, the space of scores becomes discrete, and each pixel is assigned at the end of our method to one of the score labels  ${\cal L} = \{l_1,\cdots,l_{b}\}$ (we use $b=30$).

 Let ${\bf X} = \{X_1,\cdots,X_n\}$ denote a random field of $n$ variables ($n$ is the number of pixels in an image), and let the domain of each variable be the set of score labels ${\cal L}$ ($|{\cal L}|=b$).
Our goal is to compute for each $X_j$ a single label $x_j\in {\cal L}$.
That is, we wish to find a labeling ${\bf x} \in {\cal L}^n$ that minimizes the standard MRF energy objective function:
\begin{align} \label{eq:MRF}
  E({\bf x})=\sum_{j=1}^{n}\Psi_u(x_j)+\lambda\sum_{\substack{j=1, \\ i\in {\cal N}(j)}}^{n}\Psi_p(x_j,x_i),
\end{align}
where ${\cal N}(j)$ is the neighborhood of pixel $j$, $\Psi_u(x_j)$ is the unary potential, and $\Psi_p(x_j,x_i)$ is the pairwise potential. 
The pairwise potential is a standard pairwise term that relies on intensity differences, given by:
\begin{align}
\label{eq:pairwiseTerm}
  \Psi_p(x_j,x_i) =\left \{
  \begin{tabular}{cl}
  $\frac{1}{\nabla I_j}$ & if  $x_j \neq x_i$ \\
  0 & otherwise \\
  \end{tabular}
\right.,  
\end{align}
where ${\nabla I_j}$ is the image gradient at the location of pixel~ $j$.
We propose a novel approach to compute the unary potential~$\Psi_u(x_j)$.

We set $\Psi_u(x_j)$ to be the function of a mixture of two distributions: ${\bf h}^{G}_j$, which is a discretization of pixel $j$'s initial score, $s_j$, and ${\bf h}^{A}_j$, which is based on the score of pixels in the image set that have similar feature vectors to pixel $j$. We use the letter G in our denotation of ${\bf h}^{G}_j$ because this term relies on geometry. It is based on all pixels along the epipolar lines associated with pixel $j$ (described in Sec.~\ref{InitialMaps}). We use the letter A in our denotation of ${\bf h}^{A}_j$  because this term relies on appearance. It is based on all pixels that are similar in appearance to pixel $j$.

\begin{figure*}[t]
\begin{center}

\def\arraystretch{1.5}
{\setlength{\tabcolsep}{3pt}
\begin{tabular}{cccc}

\includegraphics[width = .23\linewidth]{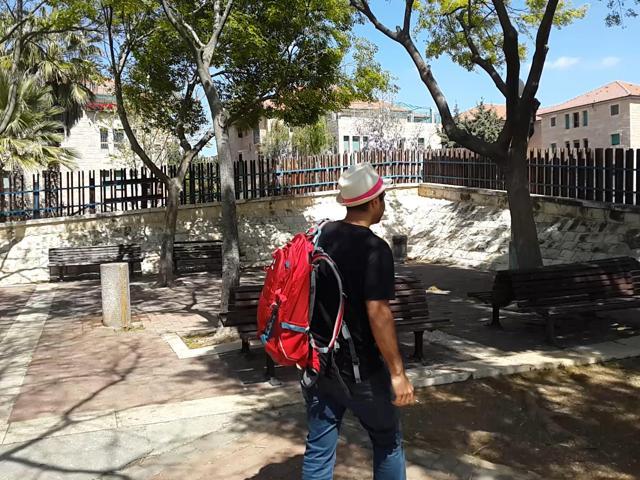}& 
\includegraphics[width = .23\linewidth]{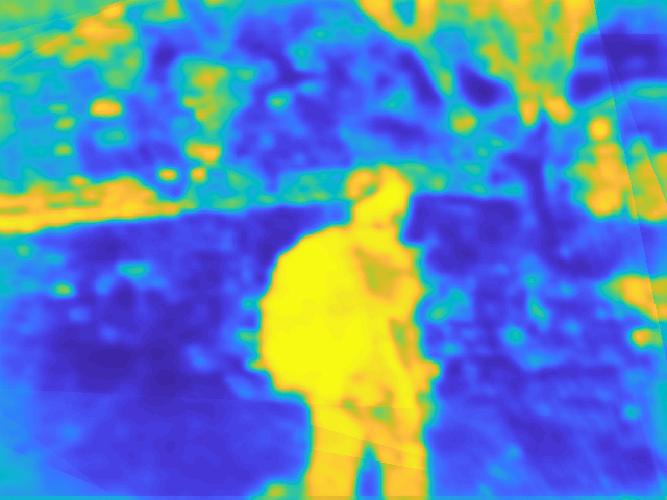}&
\includegraphics[width = .23\linewidth]{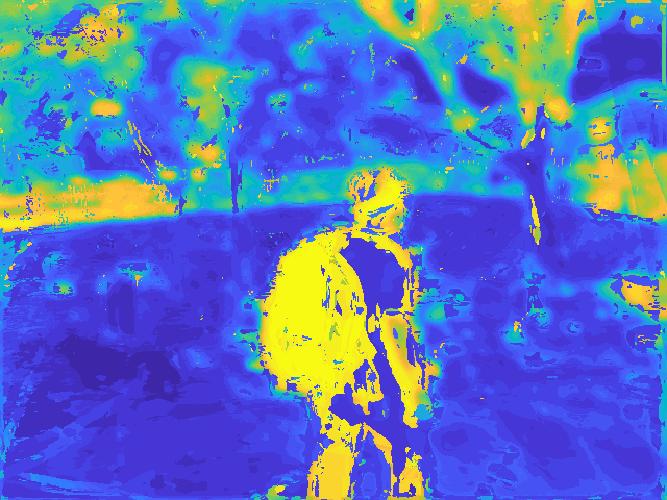} & 
\includegraphics[width = .23\linewidth]{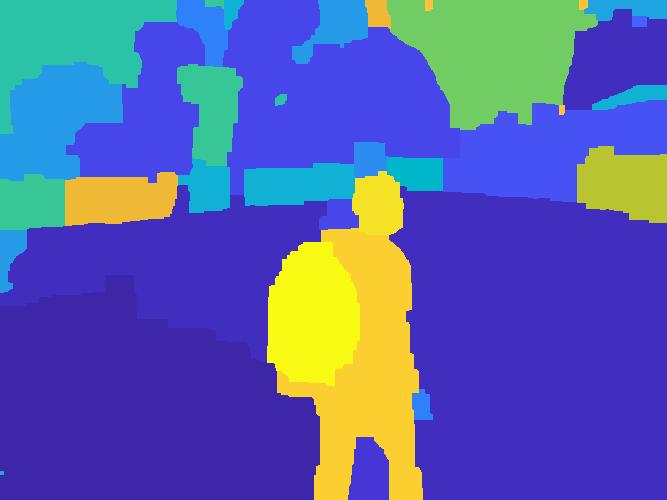}\\ 
\includegraphics[width = .23\linewidth]{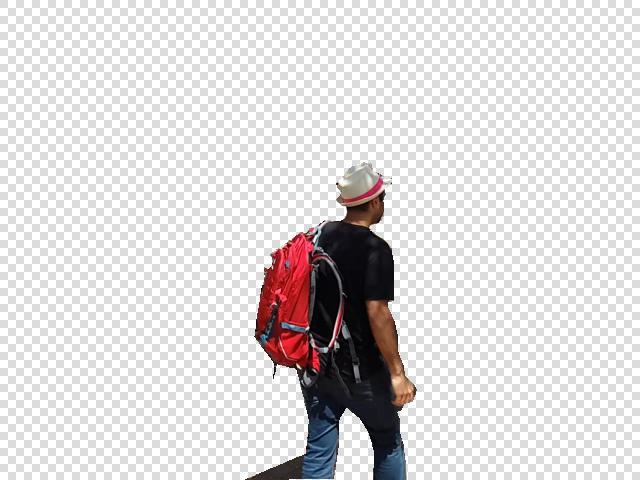} & 
\includegraphics[width = .23\linewidth]{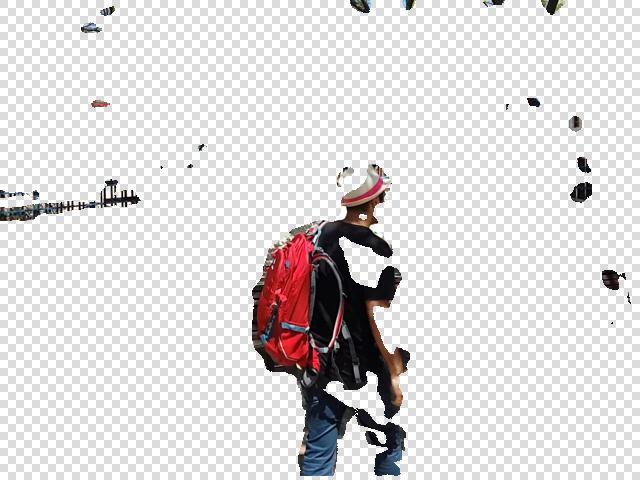} & 
\includegraphics[width = .23\linewidth]{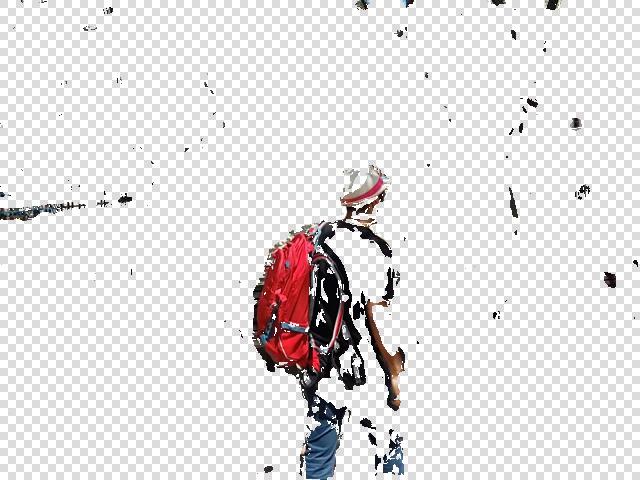}&
\includegraphics[width = .23\linewidth]{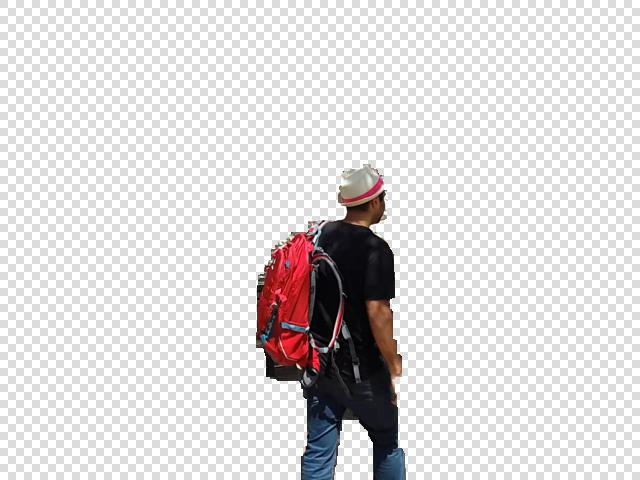}\\

\includegraphics[width = .23\linewidth]{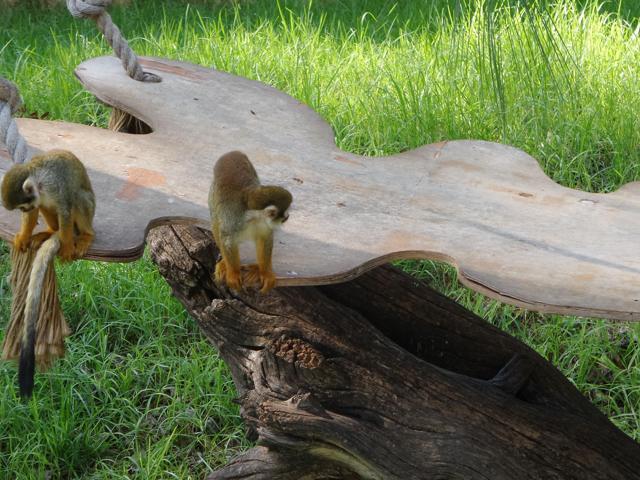}& 
\includegraphics[width = .23\linewidth]{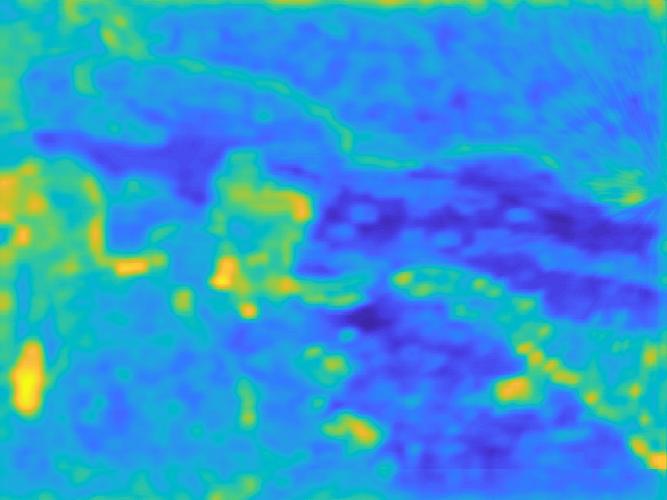}&
\includegraphics[width = .23\linewidth]{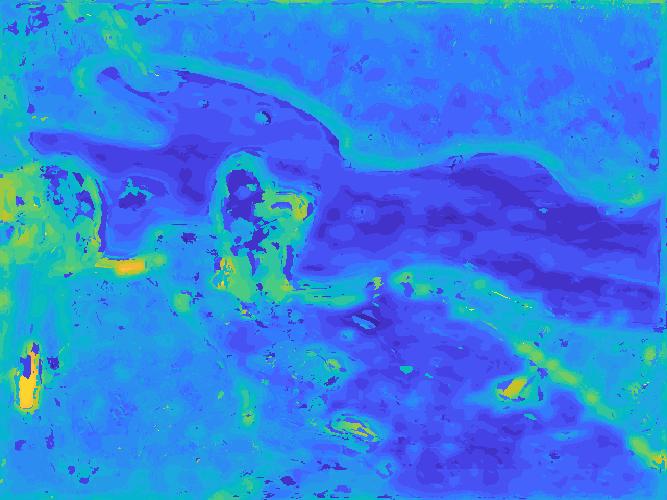} & 
\includegraphics[width = .23\linewidth]{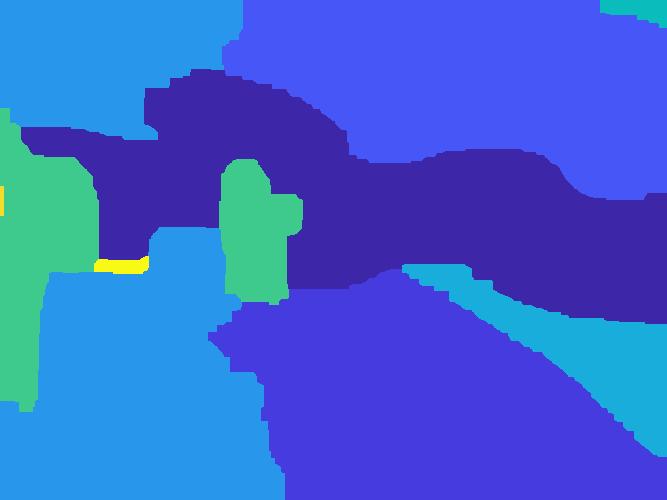}\\ 
\includegraphics[width = .23\linewidth]{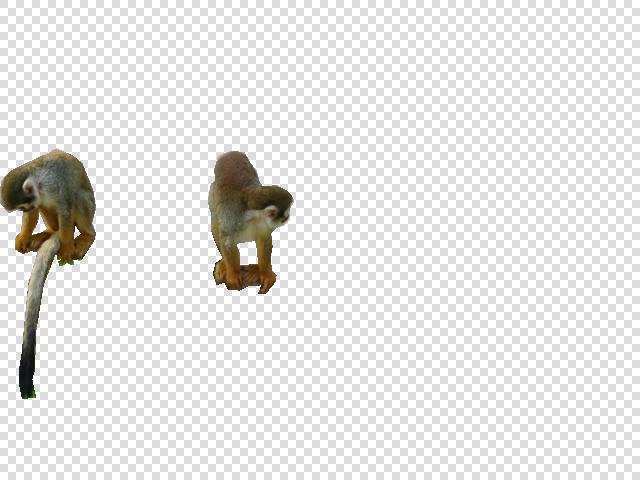} & 
\includegraphics[width = .23\linewidth]{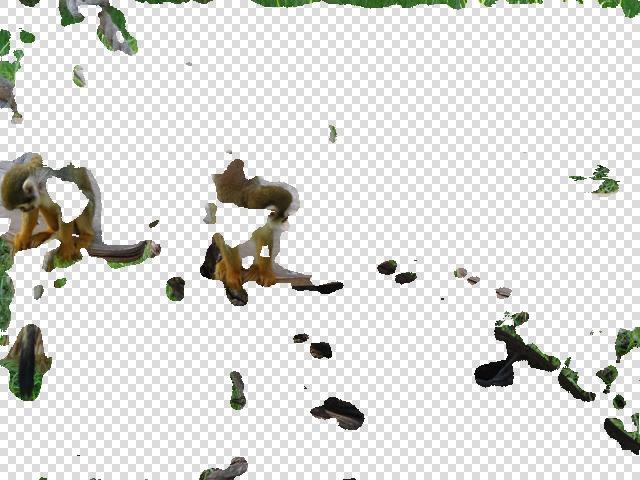} & 
\includegraphics[width = .23\linewidth]{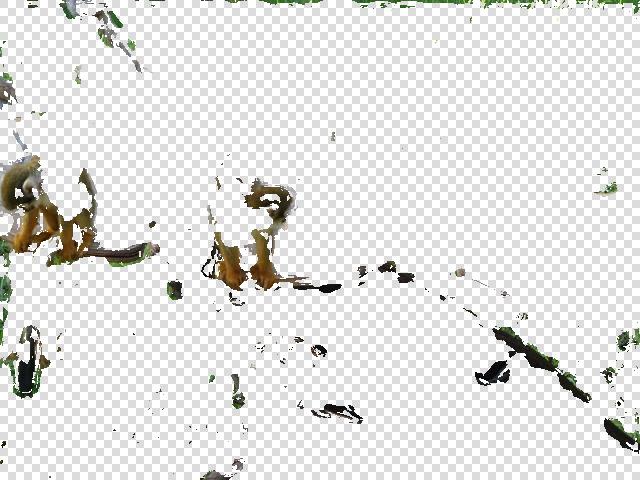}&
\includegraphics[width = .23\linewidth]{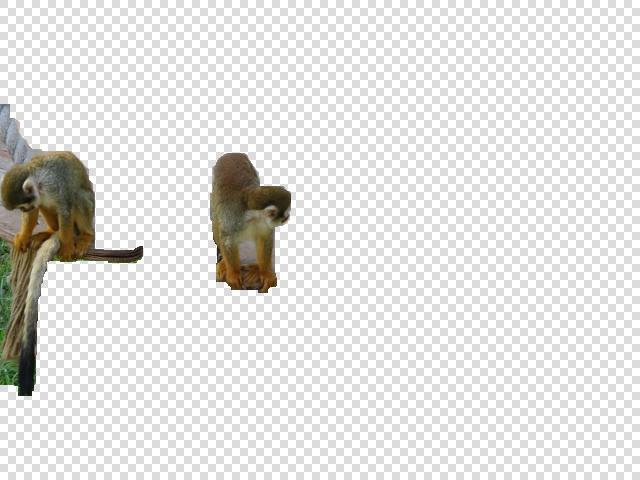}\\

(a) & (b) & (c) & (d) \\

\end{tabular}}
\caption{(a) Image and ground-truth segmentation. (b) The input to our algorithm: score maps and the segmentation based on these maps as obtained by \cite{dafni2017detecting}. (c) Results using ML estimator, i.e. using only the unary term. (d) Our results.}
\label{fig:stagesEffect}
\end{center}
\end{figure*}

The distribution ${\bf h}^{G}_j$ of labels for pixel $j$ is a delta function of the discretized initial score $s_j$ of that pixel:
\begin{align} 
\label{eq:pOriginal}
h^{G}_j(l_m)=p(x_j=l_m)=\left \{
  \begin{tabular}{cl}
  1 & if $m=\lceil s_j \cdot b \rceil$\\
  0 & otherwise \\
  \end{tabular}
\right. .
\end{align}

The distribution ${\bf h}^{A}_j$ should ideally be a function of all the pixels in all images.
Let ${\bf f}_j$ denote the feature vector of pixel~$j$ (e.g., HoG feature, a deep feature, etc.). Then our goal is to propagate the initial geometric score $s_j$ to all pixels with a similar feature vector. In practice, we first cluster all pixels' features $\{ {\bf f}_i \}_{i=1}^T$ into $K$ clusters using $K$-means (we use $K=600$), where $T$ is the total number of pixels in all the images. Then, we use only pixels that belong to the same cluster as pixel $j$ to compute ${\bf h}^{A}_j$:
\begin{eqnarray}
\label{eq:pCenter}
{\bf h}^{A}_{j} = \frac{\sum_{i=1}^N w_{i}{\bf h}^{G}_i}{\sum_{i=1}^N w_{i}},
\end{eqnarray}
where $N$ is the number of pixels assigned to that cluster and
\begin{align}
  w_{i} = e^{- \beta \frac{d_{i}^2}{M^2}}.
\end{align}
The weight $w_i$ is based on the Euclidean distance in feature space between ${\bf f}_i$ and the mean feature vector of all pixels in the cluster. $M$ is the median of the distances of all features in the cluster to the cluster center. We set $\beta = 0.3$ in all our experiments. This function is chosen such that all features affect the cluster \textit{pdf} while features far from the cluster center contribute less. Note that the pixels assigned to a cluster can come from different images and are not limited to a single image.
Combining the two distributions, we have,
\begin{align} 
\label{eq:pFinal}
	{\bf h}_{j}=\alpha {\bf h}^{G}_{j}+(1-\alpha){\bf h}^{A}_{j}.
\end{align} 
Now, we set the unary term to be:
\begin{align}
\label{eq:unaryTerm}
  \Psi_u(x_j=l_m) = -\log{h_j}(l_m).
\end{align}
The parameter $\alpha \in [0, 1]$ is the mixture weight. The effect of $\alpha$ is shown in Fig.~\ref{fig:paramsEffect}(a) and discussed in Sec.~\ref{paramsEffect}. 

An example of the computation of the probability distribution function is presented in Fig.~\ref{fig:figPosteriorProb}. It shows two different images from the same set with pixels $i$ and $j$ marked on them. Both pixels belong to the same cluster, and both depict a static 3D scene point of the same surface, so their score should be low. In practice, however, their scores are quite high (one has a score above $0.4$ and the other has a score above $0.8$). The appearance-based distribution of their cluster indicates that most pixels in that cluster are more likely to be static (i.e., have a low score). The unary term of each pixel is taken to be a mixture of both geometry-based and appearance-based pixel statistics. The final result of the MRF shows that the score of the top pixel drops dramatically from over $0.8$ to about $0.2$, while the score of the bottom pixel does not change much. This shows the importance of using  both appearance and proximity.

The output of the MRF is a new set of score maps for each image ${\bf \tilde s_r}$ that significantly improves the  dynamic object segmentation over the initial map ${\bf  s_r}$, as described in the next section.

As for the implementation, to solve the MRF energy minimization equation we used the graph-cut solver published by~\cite{Boykov2004,Boykov2001,Kolmogorov2004}. 

\begin{figure*}
\begin{center}

\def\arraystretch{1.5}
{\setlength{\tabcolsep}{3pt}

\begin{tabular}{llllll}

\includegraphics[width = .154\linewidth]{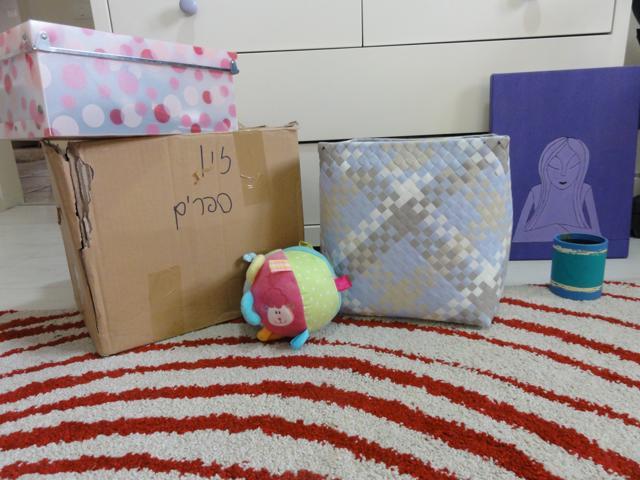} & 
\includegraphics[width = .154\linewidth]{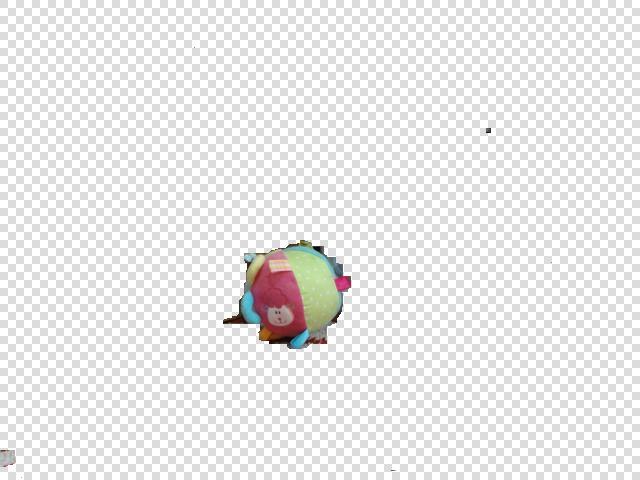} & 
\includegraphics[width = .154\linewidth]{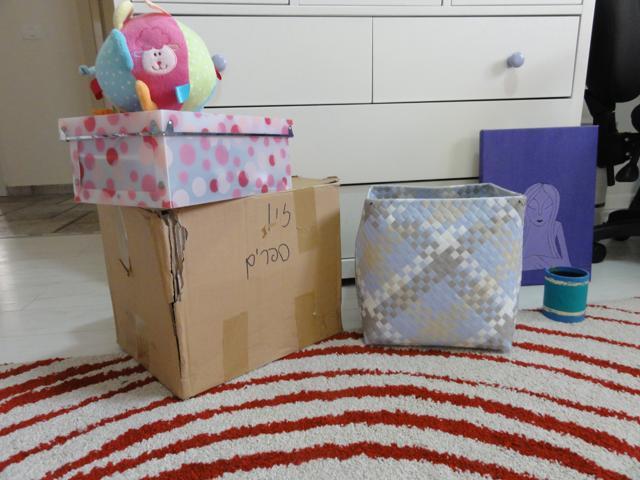} & 
\includegraphics[width = .154\linewidth]{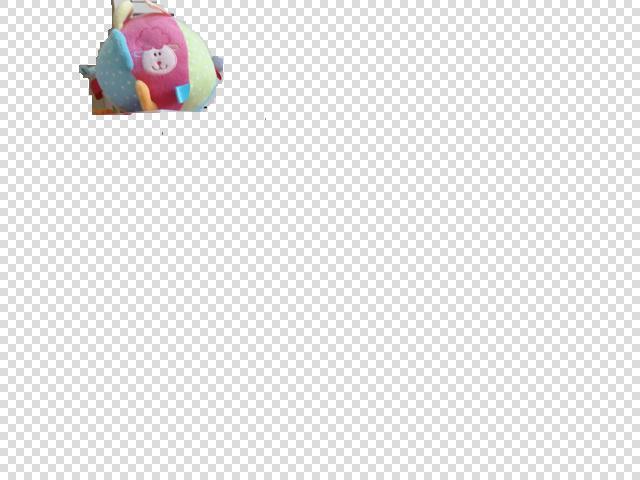} &
\includegraphics[width = .154\linewidth]{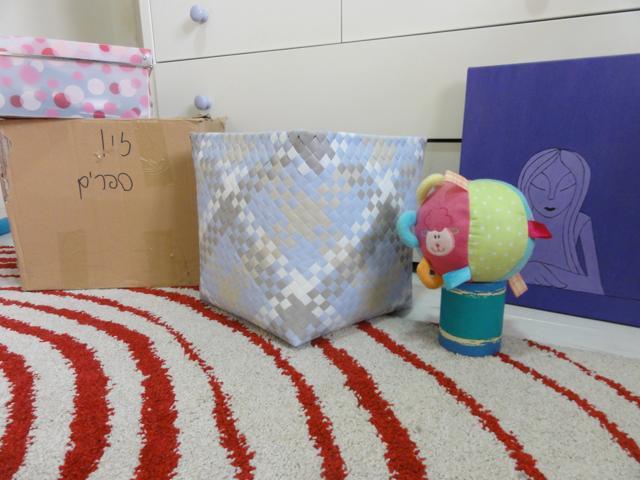} & 
\includegraphics[width = .154\linewidth]{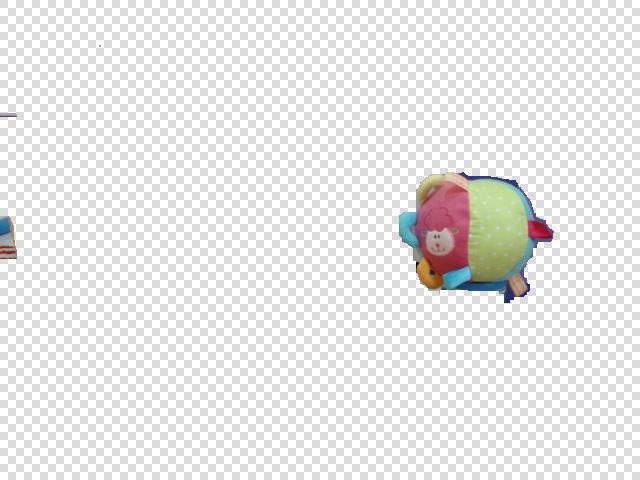}\\

\includegraphics[width = .154\linewidth]{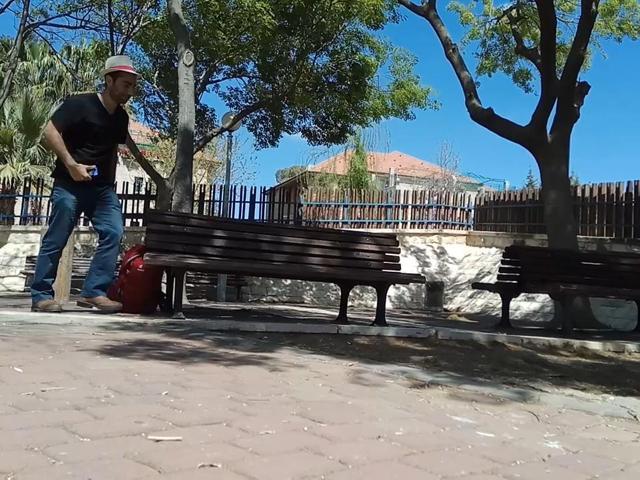}& 
\includegraphics[width = .154\linewidth]{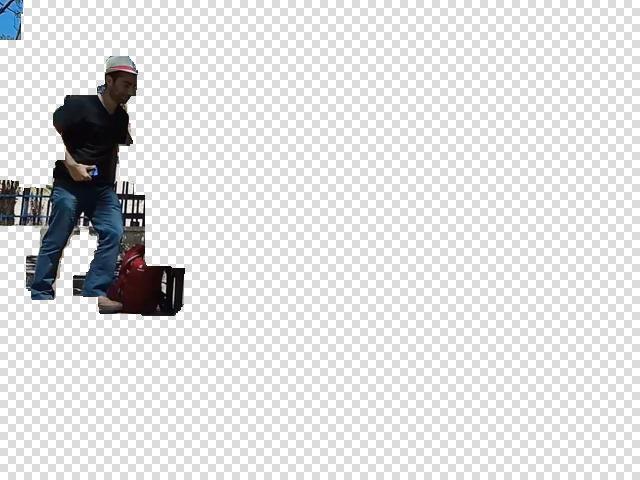}& 
\includegraphics[width = .154\linewidth]{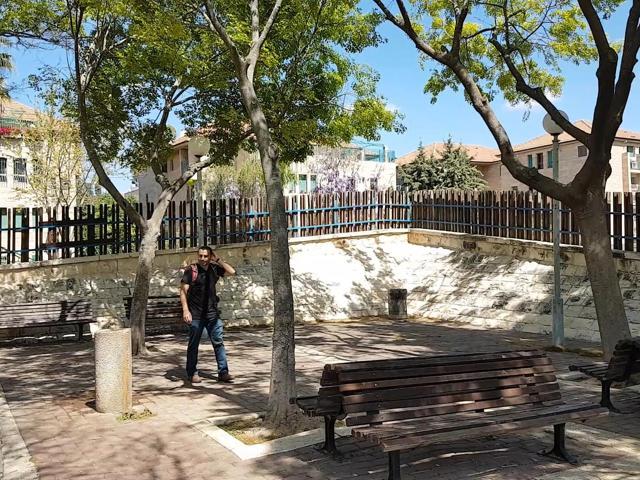}& 
\includegraphics[width = .154\linewidth]{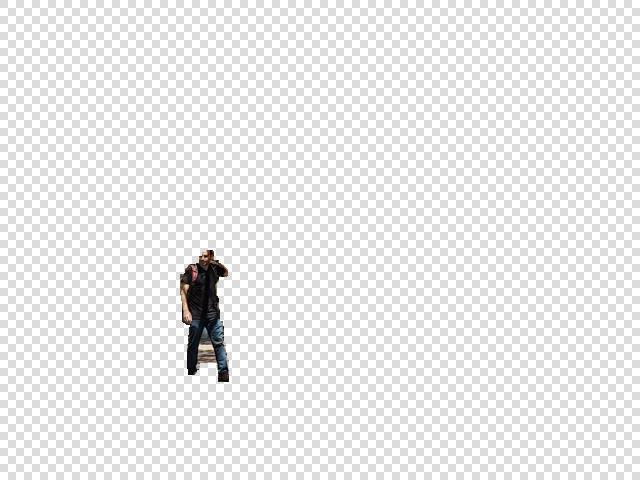}& 
\includegraphics[width = .155\linewidth]{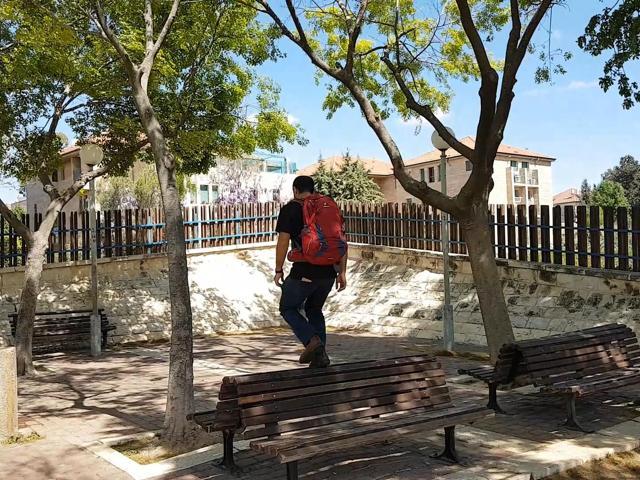}& 
\includegraphics[width = .154\linewidth]{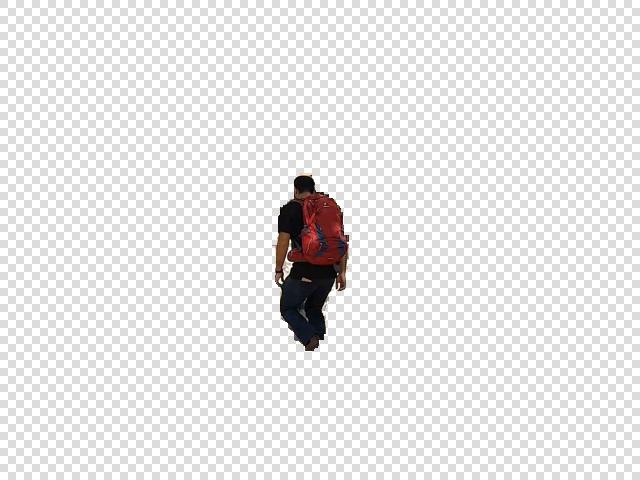}\\

\includegraphics[width = .154\linewidth]{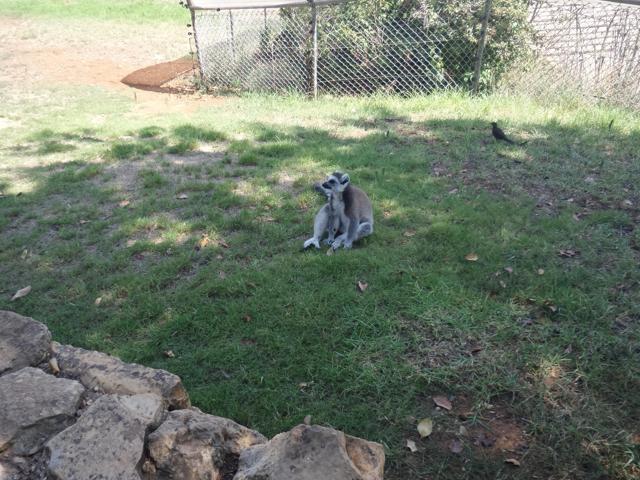} & 
\includegraphics[width = .154\linewidth]{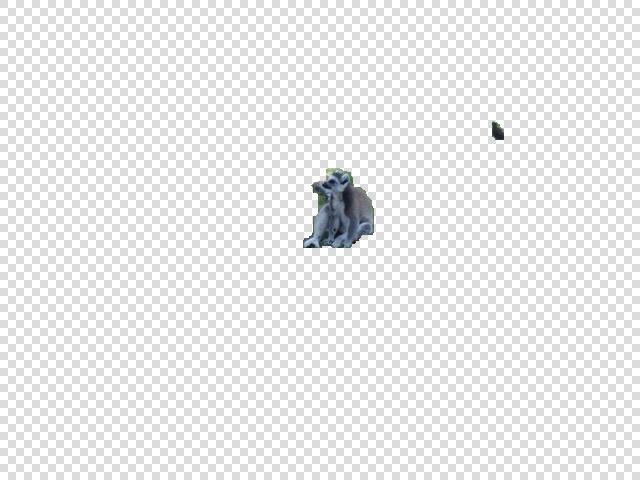} & 
\includegraphics[width = .154\linewidth]{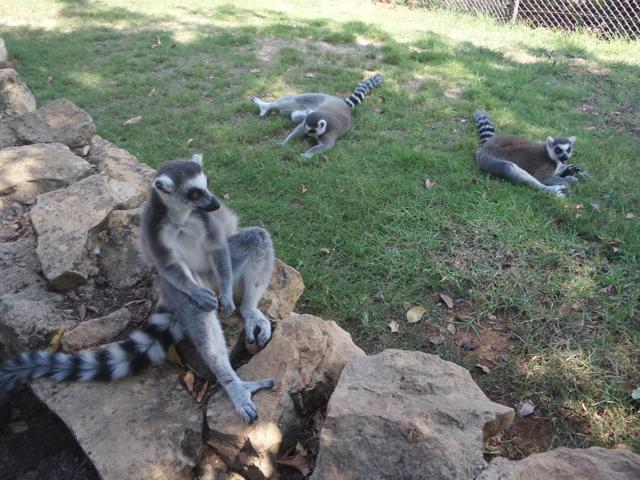} & 
\includegraphics[width = .154\linewidth]{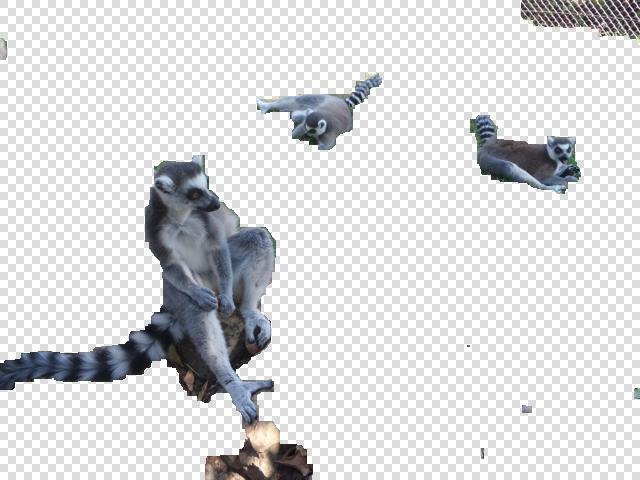} & 
\includegraphics[width = .154\linewidth]{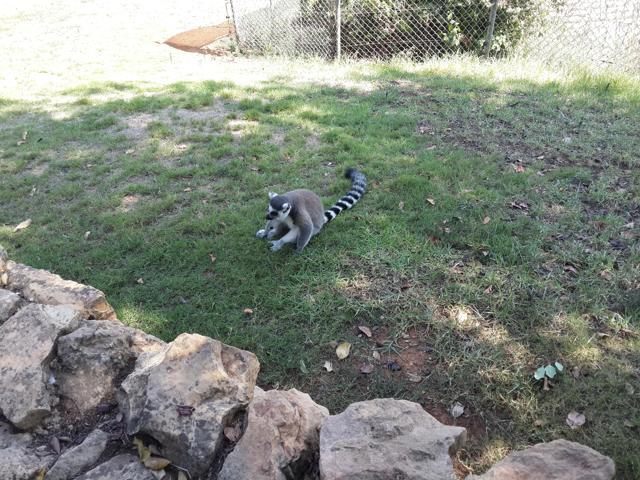} & 
\includegraphics[width = .154\linewidth]{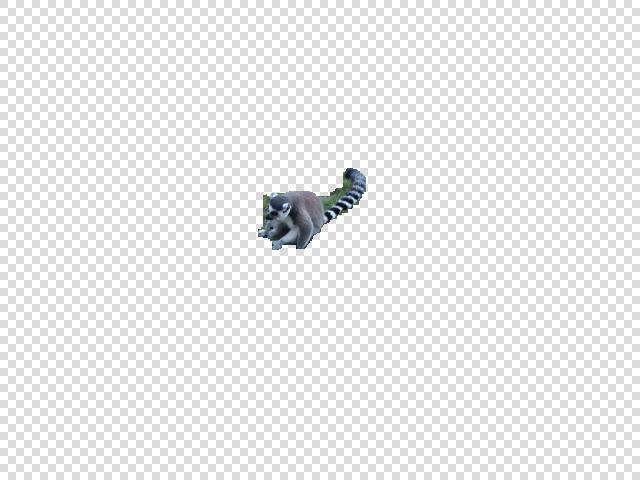} \\

\includegraphics[width = .154\linewidth]{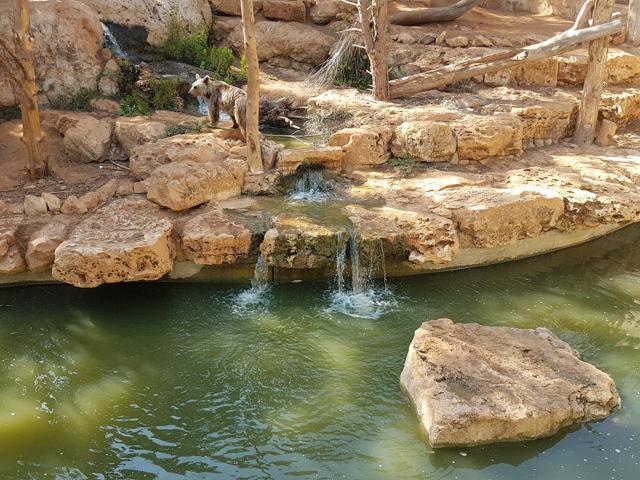} &
\includegraphics[width = .154\linewidth]{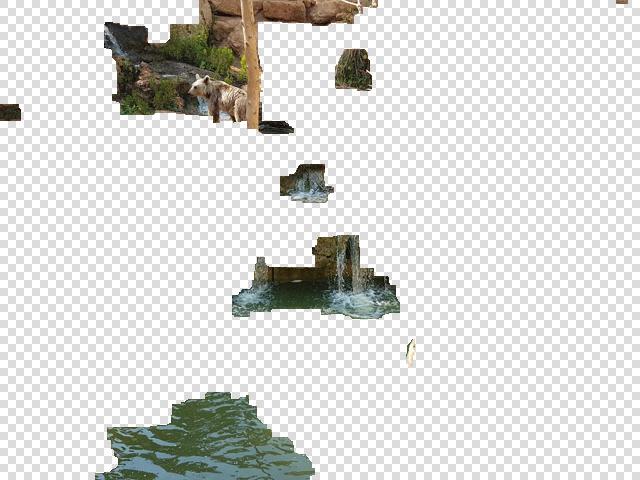} & 
\includegraphics[width = .155\linewidth]{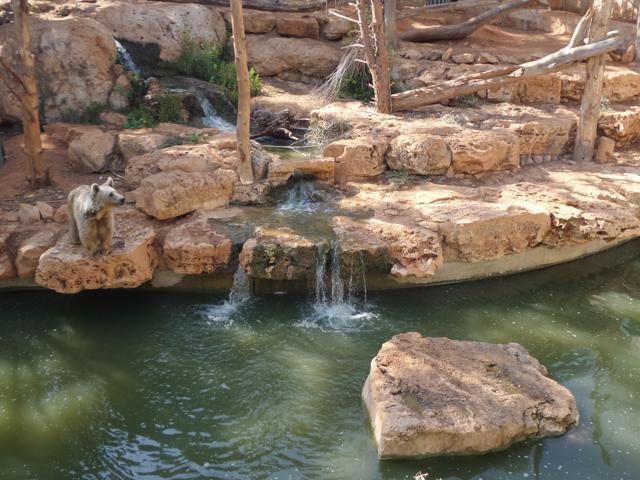} & 
\includegraphics[width = .154\linewidth]{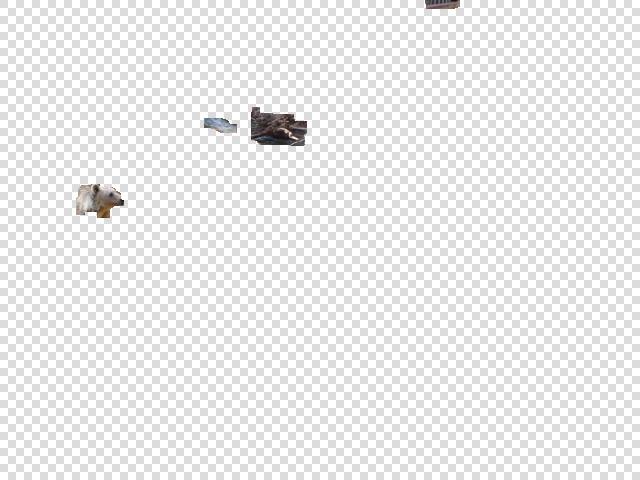} &  
\includegraphics[width = .154\linewidth]{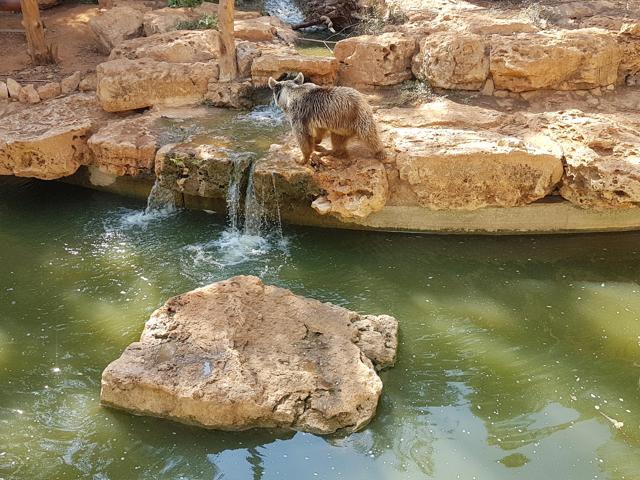} &
\includegraphics[width = .154\linewidth]{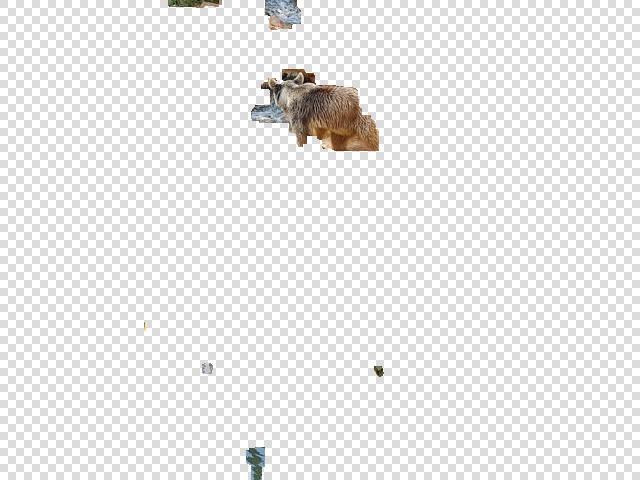} \\

\end{tabular}}
\caption{Example of CrowCam image-sets together with a sample of our segmentation. The first row shows the "Toy-ball" scene from the previous dataset by \cite{dafni2017detecting}. The second to fourth rows, "Jlm1", "Zoo4" and "Zoo8" respectively, are new scenes we captured}
\label{fig:crowdCamImages}
\end{center}
\end{figure*}

\section{Experiments} \label{Experiments}
   
We implemented our method using MATLAB.
 We tested our method on the dataset used by~ \cite{dafni2017detecting} as well as on a new dataset we captured. The results of our method are  compared  to the results of the only two methods~\cite{dafni2017detecting,Kanojia2018} that address the task of dynamic object segmentation from CrowdCam data. We also compare our results with ML estimator (see Sec.~\ref{algVariants}), and with DenseCRF~\cite{krahenbuhl2011efficient}. 


\vspace{0.2cm}
\noindent{\bf Data Sets:~}
The images in each set were captured by several cameras, at a different timing from different viewpoints, as typical for CrowdCam images. 
The dataset used by~\cite{dafni2017detecting} contains six scenes, with one or more  dynamic objects such as people, bicycles and toy balls. The images were taken by Dafni~\etal~\cite{dafni2017detecting}, Park~\etal~\cite{park20103d}, and Basha~\etal~\cite{basha2012photo}. Each set consists of four to ten images.
We collected  a new dataset to extend the variety of considered scenes. Our new dataset contains~$14$ scenes, with dynamic objects such as people and animals. 
Each set consists between five to ten images.
Examples of scenes are presented in Fig.~\ref{fig:crowdCamImages}, and in the supplementary material. We used 20 scenes in our experiments.


\noindent{\bf Evaluation:~}
The  segmented  ground  truth masks of  dynamic objects were provided by~\cite{dafni2017detecting} and for the new dataset, they were manually marked. In~\cite{dafni2017detecting} a few scenes (Climbing, Skateboard and Playground) are marked for both \textit{`ground truth'} (of the dynamic objects) and \textit{`don't care'} regions, which are disregarded when evaluating the results.
The Jaccard measure, also known as Intersection over Union (IoU), is a standard measure for binary map segmentation results.  It is used to compare the computed and the ground-truth region.

The output of our algorithm, as well as the discussed algorithms (CRF, ML and Dafni~\etal~\cite{dafni2017detecting}), is a score map for each image rather than a binary map. Inspired by the evaluation methodology of the Berkeley Segmentation Dataset~\cite{MartinFTM01} and~\cite{dafni2017detecting}, we obtain a binary map by applying a threshold operation on the score maps and then use the Jaccard measure. We present the results of a {\em per-image} threshold, which is chosen independently for each image such that it maximizes the Jaccard measure for each image. We also present the results obtained by using a {\em per-set} threshold, where the same threshold is chosen for the entire set, such that it maximizes the average Jaccard measure of all images in a given set.

\begin{table*}
\begin{center}
\small
\begin{tabular}{|l|c|c|c|c||c|c|c|}
\hline
Scene~~(\#images) & Dafni~\cite{dafni2017detecting} & DafniVGG & Kanojia~\cite{Kanojia2018} & DCRF~\cite{krahenbuhl2011efficient} & ML & CDRSs & CDRSm \\
\hline\hline
Helmet~~(4)     & 0.53~\bsls~0.36 & 0.69~\bsls~0.65 & -            & 0.69~\bsls~0.66          & 0.66~\bsls~0.62 & 0.77~\bsls~0.77                   & \textbf{0.83}~\bsls~\textbf{0.83}  \\
Climbing~~(10)  & 0.15~\bsls~0.13 & 0.40~\bsls~0.36 & -~\bsls~0.34 & 0.42~\bsls~0.38          & 0.30~\bsls~0.28 & \textbf{0.59}~\bsls~0.52          & \textbf{0.59}~\bsls~\textbf{0.59}  \\
Skateboard~~(5) & 0.44~\bsls~0.42 & 0.44~\bsls~0.42 & -~\bsls~0.50 & 0.49~\bsls~0.42          & 0.42~\bsls~0.39 & \textbf{0.71}~\bsls~0.61          & 0.67~\bsls~\textbf{0.67} \\
Toy-ball~~(7)   & 0.63~\bsls~0.60 & 0.60~\bsls~0.57 & -~\bsls~0.44 & 0.66~\bsls~0.60          & 0.54~\bsls~0.52 & 0.74~\bsls~\textbf{0.66}          & \textbf{0.77}~\bsls~0.64  \\
Playground~~(7) & 0.37~\bsls~0.32 & 0.40~\bsls~0.33 & -~\bsls~0.37 & 0.41~\bsls~0.30          & 0.39~\bsls~0.35 & \textbf{0.50}~\bsls~0.41          & 0.48~\bsls~\textbf{0.48} \\
Basketball~~(8) & 0.48~\bsls~0.47 & 0.48~\bsls~0.46 & -~\bsls~0.51 & \textbf{0.64}~\bsls~0.60 & 0.48~\bsls~0.46 & 0.59~\bsls~0.54                   & 0.63~\bsls~\textbf{0.61}  \\
Jlm1~~(8)       & 0.48~\bsls~0.45 & 0.57~\bsls~0.51 & -            & 0.61~\bsls~0.60          & 0.48~\bsls~0.44 & 0.69~\bsls~0.66                   & \textbf{0.72}~\bsls~\textbf{0.70}  \\
Jlm2~~(8)       & 0.60~\bsls~0.57 & 0.60~\bsls~0.56 & -            & 0.74~\bsls~0.64          & 0.59~\bsls~0.56 & 0.78~\bsls~0.74                   & \textbf{0.84}~\bsls~\textbf{0.82} \\
Chess~~(5)      & 0.14~\bsls~0.11 & 0.24~\bsls~0.24 & -            & 0.34~\bsls~0.32          & 0.33~\bsls~0.32 & 0.33~\bsls~0.33                   & \textbf{0.37}~\bsls~\textbf{0.36} \\
Shelf~~(8)      & 0.25~\bsls~0.18 & 0.44~\bsls~0.27 & -            & 0.53~\bsls~0.36          & 0.40~\bsls~0.28 & \textbf{0.61}~\bsls~0.37          & 0.58~\bsls~\textbf{0.48} \\
Zoo1~~(7)       & 0.21~\bsls~0.16 & 0.31~\bsls~0.28 & -            & 0.32~\bsls~0.26          & 0.28~\bsls~0.24 & \textbf{0.37}~\bsls~\textbf{0.35} & 0.36~\bsls~\textbf{0.35} \\
Zoo2~~(8)       & 0.06~\bsls~0.04 & 0.12~\bsls~0.10 & -            & 0.11~\bsls~0.09          & 0.09~\bsls~0.07 & \textbf{0.24}~\bsls~\textbf{0.15} & 0.17~\bsls~0.11 \\
Zoo3~~(6)       & 0.10~\bsls~0.07 & 0.29~\bsls~0.27 & -            & 0.47~\bsls~0.42          & 0.32~\bsls~0.30 & 0.60~\bsls~0.52                   & \textbf{0.68}~\bsls~\textbf{0.63} \\
Zoo4~~(10)      & 0.37~\bsls~0.26 & 0.46~\bsls~0.35 & -            & \textbf{0.70}~\bsls~0.51 & 0.44~\bsls~0.33 & 0.60~\bsls~0.42                   & 0.65~\bsls~\textbf{0.52} \\
Zoo5~~(7)       & 0.19~\bsls~0.16 & 0.33~\bsls~0.31 & -            & 0.40~\bsls~0.34          & 0.35~\bsls~0.33 & 0.45~\bsls~0.38                   & \textbf{0.48}~\bsls~\textbf{0.46} \\
Zoo6~~(9)       & 0.06~\bsls~0.05 & 0.13~\bsls~0.10 & -            & 0.25~\bsls~0.17          & 0.15~\bsls~0.09 & \textbf{0.29}~\bsls~\textbf{0.20} & 0.22~\bsls~\textbf{0.20} \\
Zoo7~~(9)       & 0.13~\bsls~0.07 & 0.17~\bsls~0.12 & -            & 0.17~\bsls~0.12          & 0.18~\bsls~0.11 & \textbf{0.27}~\bsls~0.14          & 0.25~\bsls~\textbf{ 0.18} \\
Zoo8~~(8)       & 0.02~\bsls~0.02 & 0.16~\bsls~0.13 & -            & 0.17~\bsls~0.14          & 0.15~\bsls~0.12 & 0.33~\bsls~0.26                   & \textbf{0.40}~\bsls~\textbf{0.32}\\
Zoo9~~(6)       & 0.19~\bsls~0.16 & 0.36~\bsls~0.33 & -            & 0.47~\bsls~0.41          & 0.36~\bsls~0.32 & 0.61~\bsls~0.53                   & \textbf{0.71}~\bsls~\textbf{0.67} \\
Zoo10~~(7)      & 0.21~\bsls~0.17 & 0.29~\bsls~0.26 & -            & 0.46~\bsls~0.41          & 0.30~\bsls~0.26 & 0.45~\bsls~0.37                   & \textbf{0.55}~\bsls~\textbf{0.43} \\

\hline
Mean        & 0.28~\bsls~0.24 & 0.37~\bsls~0.33 & -    & 0.45~\bsls~0.39 & 0.36~\bsls~0.32 & 0.53~\bsls~0.45 & \textbf{0.55}~\bsls~\textbf{0.50} \\
\hline
\end{tabular}
\end{center}
\caption{Results on CrowdCam datasets in terms of Jaccard index. 
The threshold on the score maps is chosen per image~\bsls~per set. ML: using only the unary term in the MRF equation. CDRSs: our algorithm using a single image when computing ${\bf h}_j$. CDRSm: our algorithm using multiple images when computing ${\bf h}_j$}
\label{tbl:resultsCombined}
\end{table*}

\subsection{Results}
\label{resultsAndComparison}

Before diving into the details, here is a summary of our findings. The current state of the art~\cite{dafni2017detecting} achieved an average Jaccard score of~$\textbf{0.28}$. By replacing the hand crafted HoG features with deep features we achieved a Jaccard score of~$0.37$. Adding proximity cues (in the form of MRF formulation) improves the Jaccard score to~$0.45$. Finally, adding appearance cues lets us achieve a Jaccard score of~$\textbf{0.55}$ which is almost double the score of the current state of the art.

Now, to a detailed analysis of the results. We present the results of our method on all datasets described above using the \textit{same} set of parameters for all image sets: $\alpha=0.2$ when calculating the final dynamic score \textit{pdf} ${\bf h}_j$ (Eq.~\ref{eq:pFinal}), $\lambda=450$ in the MRF energy objective function (Eq.~\ref{eq:MRF}), and $K=600$ as the number of clusters when calculating ${\bf h}_j^A$ (Eq.~\ref{eq:pCenter}). 
Two levels of sharing appearance information were considered. The first is when ${\bf h}^{A}_{j}$ is computed from a single image (CDRSs), and the second is when it is computed from multiple  images in the set  (CDRSm). 

Qualitative results are shown in Fig.~\ref{fig:stagesEffect} and in Fig.~\ref{fig:crowdCamImages}. Fig.~\ref{fig:stagesEffect} illustrates the strength of our method, as a much better segmentation is obtained using the calculated score maps. The score maps calculated by our method are  more accurate thanks to our algorithm's sharing of statistics between neighboring pixels in both appearance and spatial domains. Fig.~\ref{fig:crowdCamImages} presents images from a few scenes that reveal the difficulties of dynamic region segmentation in CrowdCam image sets. The top two rows show that a good segmentation is obtained even when dynamic objects appear different for various reasons. This includes changes in viewpoint, differences in color, changes in illumination, and the use of different cameras. Multiple objects are segmented as our method gives no importance to the number of dynamic objects to segment. This is clearly seen in the segmentation of ``Zoo4" where all monkeys are correctly segmented, while a small portion of the background is erroneously segmented as a moving region as-well. The last row presents a very challenging scene of a moving bear (``Zoo8"). The results for this scene are not as good because the bear has a similar appearance in texture and color to its background. The initial score maps are very inaccurate as a results, and thus our method fails on this scene.

\begin{figure*}
\begin{center}

\begin{tabular}{ccc}

\includegraphics[width = .30\linewidth]{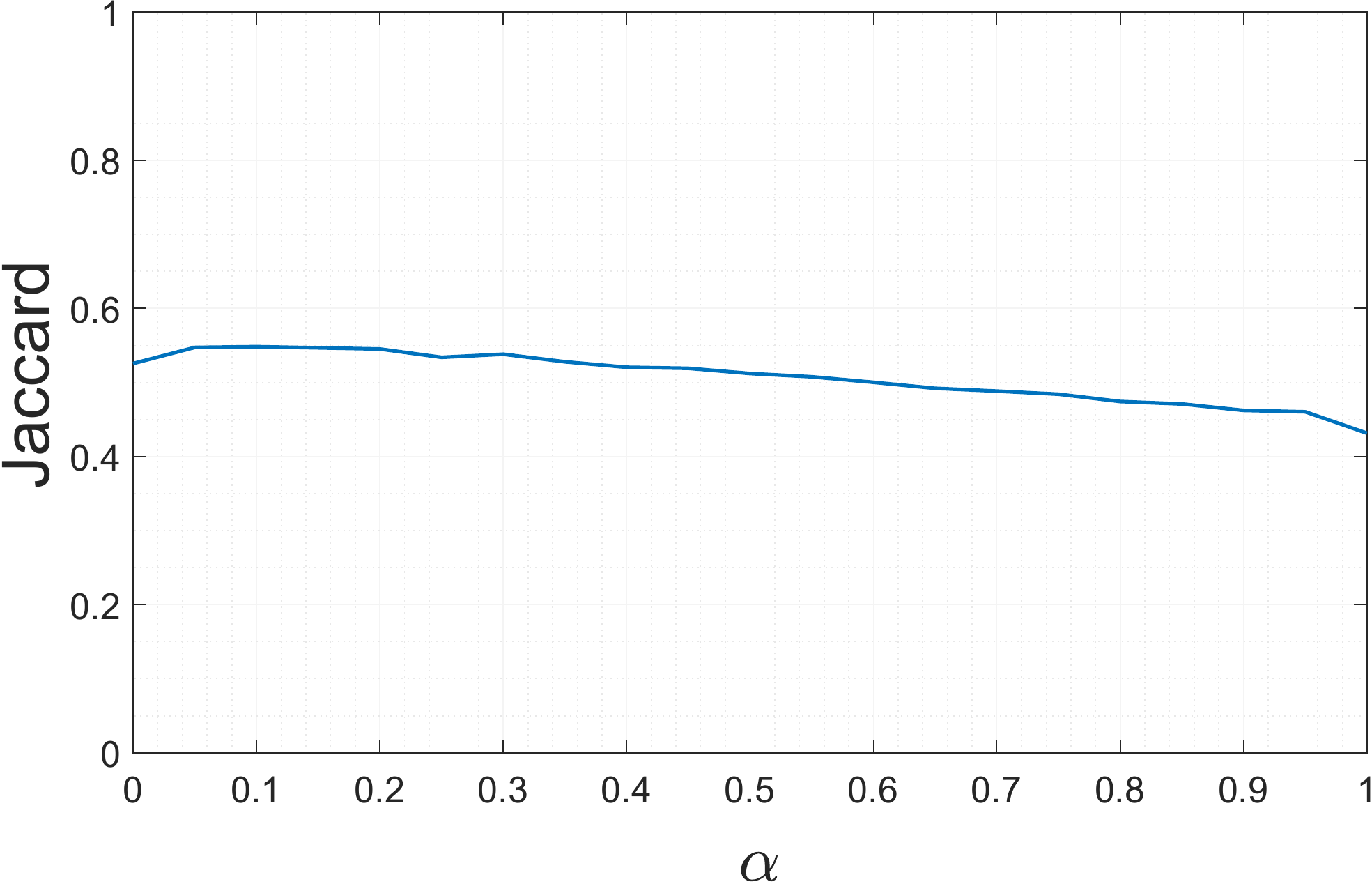} &
\includegraphics[width = .30\linewidth]{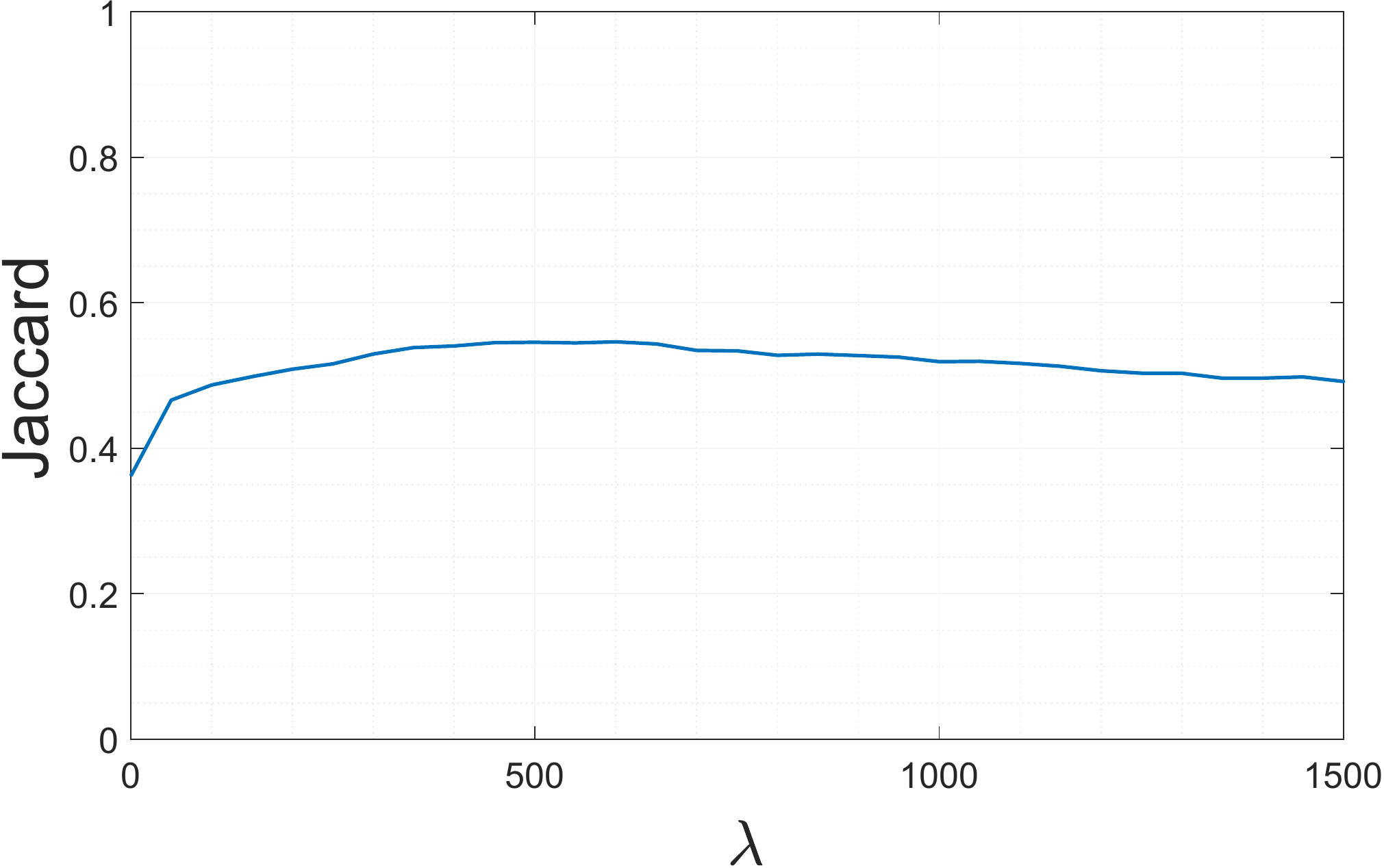} &
\includegraphics[width = .30\linewidth]{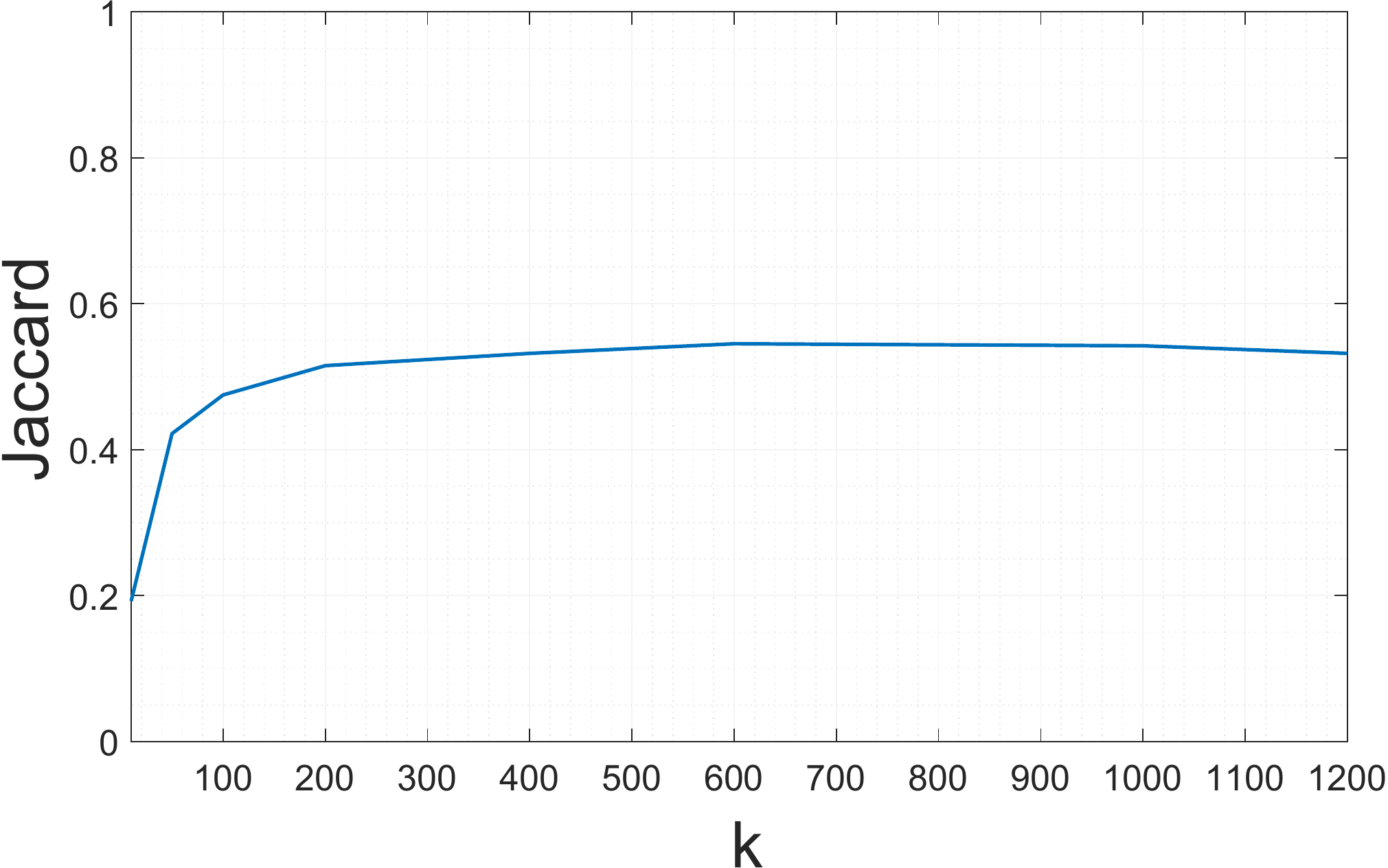} \\

(a) & (b) & (c)

\end{tabular}
\caption{Mean Jaccard index over image sets using different parameters in our algorithm. The results were calculated using multiple images to calculate $h^{A}_j$ (CDRSm) (a) Changing the value of $\alpha$ in equation~\ref{eq:pFinal}. (b) Changing the value of $\lambda$ in equation~\ref{eq:MRF}. (c) Changing the number of centers when calculating K-means.}
\label{fig:paramsEffect}
\end{center}
\end{figure*}

The quantitative results using the per-image and per-set thresholds are summarized in Table~\ref{tbl:resultsCombined}.
The results of using multiple images, CDRSm, are better than those obtained by  using a single image, CDRSs. Interestingly, while CDRSm performs better by a slight margin when the per-image threshold is used, it performs much better than the other methods when the per-set threshold is used. 
This is due to the information sharing between images, which results in the same score for similar regions across images (e.g., a grass field that is presented in all images of the scene get a similar score in the score maps). Hence, the same threshold will apply.

We  compare our results with the results obtained by the only two algorithms designed to solve the  same task considered in the paper~\cite{dafni2017detecting,Kanojia2018}.  It is apparent from several examples in Fig.~\ref{fig:stagesEffect} that the score maps produced by our method, as well as the final segmentation, improve over the initial score map computed by~\cite{dafni2017detecting}.
For the qualitative comparison, we present in Table~\ref{tbl:resultsCombined} the  reported results of the algorithm used in~\cite{dafni2017detecting} and~\cite{Kanojia2018} for the dataset in~\cite{dafni2017detecting}, and we refer to these algorithms {\em Dafni} and {\em Kanojia}, respectively. 
 We also modified the original implementation of~\cite{dafni2017detecting}, by using 
  VGGs as features (instead of HoGs) ({\em DafniVGG}), as described in Sec.~\ref{InitialMaps}.
 Note that~\cite{Kanojia2018} reported only the per-set threshold results. 
The results show that using VGG (DafniVGG) improves the original results of~\cite{dafni2017detecting} by~$34$ percent in the Jaccard measure. Our method outperforms the existing methods for CrowdCam dynamic object segmentation,~\cite{dafni2017detecting,Kanojia2018}. On average, we improved the results of~\cite{dafni2017detecting} by~$50$ percent in the Jaccard measure after the change in features (DafniVGG) and outperformed the results of~\cite{Kanojia2018} by more than $38$ percent in the Jaccard measure.

\noindent{\bf Comparison to DenseCRF:~}
\label{CRF_comparison}
We compare our results to the DenseCRF~\cite{krahenbuhl2011efficient} algorithm, that computes a label per pixel given a \textit{pdf} of each pixel. The DenseCRF algorithm is somewhat similar to our algorithm in the sense that both search for the correct label per pixel, using the pixel's neighbors in both spatial and feature spaces. The main difference is that the DenseCRF incorporates both spatial and feature neighborhood constraints in the pairwise potential of the energy minimization problem, while we incorporate the feature neighborhood constraint in the unary potential. 

Notice that we use deep-features when computing $h_j^A$ (as part of the unary term calculation) while~\cite{krahenbuhl2011efficient} use RGB values in the energy minimization pairwise term.  
Another important difference is that we use information from all images in a set, whereas the DenseCRF considers pixels from the image itself while assigning greater importance to closer pixels.

In practice, the DenseCRF uses a \textit{pdf} as input rather than a single score. Hence,  we smoothed the score values  by {\em DafniVGG} and used it as a \textit{pdf} input. The results are summarized in Table~\ref{tbl:resultsCombined}. Using threshold on the score maps, we present better segmentation results, with an average Jaccard score difference of 0.07 in comparison to CDRSs and 0.1 in comparison to CDRSm.

As for the implementatoin, using the code published by~\cite{krahenbuhl2011efficient}, we noticed that every change in DenseCRF parameters resulted in improving the results of some scenes while having the opposite effect on others. We performed grid search to find the parameters that achieve the best overall Jaccard results for the DenseCRF method. 

\subsection{Algorithm Variants}
\label{algVariants}

We also tested the effect of using MRF on the computed \textit{pdf}, by applying an ML estimator instead. That is,  the final score of each pixel is the label with the maximal value of the \textit{pdf} and  proximity is not used.  This is equivalent to setting $\lambda=0$ in Eq.~\ref{eq:MRF} (i.e., only the unary term is used without the pairwise term).  The quantitative differences are shown in Fig.~\ref{fig:stagesEffect}, and the qualitative ones in Table~\ref{tbl:resultsCombined}. It is apparent that the use of MRF is crucial for the performance of the method as it removes noise and enforces spatial smoothness. The results when using the ML estimator drop by 0.18 on average, in comparison to CDRSm.

\subsection{Effect of Parameters}
\label{paramsEffect}
In this section we test the effect of our algorithm parameters, $\alpha$ (Fig.~\ref{fig:paramsEffect}(a)), $\lambda$ (Fig.~\ref{fig:paramsEffect}(b)), and $K$ of the $K-means$ algorithm
(Fig.~\ref{fig:paramsEffect}(c)).

\noindent{$\bm{ \alpha}$:~}
The mixture weight from Eq.~\ref{eq:pFinal} determines the extent to which the geometric and appearance cues affect the pixels' \textit{pdf}. Setting $\alpha=1$ causes the pixels' \textit{pdf} to depend solely on the geometry based score, ignoring global information. Setting $\alpha=0$ has the opposite effect: each pixel loses its local initial score. It is clear from Fig.~\ref{fig:paramsEffect}(a) that setting $\alpha$ closer to $1$ yields poorer results. This implies that the global appearance \textit{pdf}, ${\bf h}^{A}_j$, is crucial for our method to work properly. 

\noindent{ $\bm{ \lambda}$:~}
The $\lambda$ parameter moderates the spatial constraint in our method. It weights the pairwise potential in the MRF energy term of Eq.~\ref{eq:MRF} such that a higher $\lambda$ value results in a smoother scoring result. 
Fig.~\ref{fig:paramsEffect}(b) illustrates the importance of the MRF solver and  $\lambda$. When $\lambda$ is set to a small value, the spatial constraint has less effect on the final labeling, causing noisy results. This can also be seen in Fig.~\ref{fig:stagesEffect}(c), where the ML estimator is obtained by setting $\lambda=0$, i.e., using only the unary term. Setting a proper value for $\lambda$ significantly improves the results, as seen in Fig.~\ref{fig:stagesEffect}(d). There is a wide range of $\lambda$ values where our algorithm works properly, as shown in Fig.~\ref{fig:paramsEffect}(b).

\noindent{\textbf{K in K-means}:~}
We tested the effect of $K$, the number of clusters in the $K$-means algorithm, on the final Jaccard measure. The robustness of our method to $K$ is shown in Fig.~\ref{fig:paramsEffect}(c). As expected, for a value greater than ${\sim }200$, the method performs well and the results do not change by much when increasing this value. 

\section{Conclusions}
We proposed a comprehensive solution to the problem of dynamic object segmentation in CrowdCam images.
Our approach combines geometry, appearance and proximity in a novel manner. We tested our method on existing datasets, as well as a new dataset collected by us. Experiments suggest our method surpass the current state-of-the-art by a huge margin. The experiments also show the contribution of each of the cues used by our method, geometry, appearance and proximity, as well as the effect of the manner in which these cues are integrated.


{\small
\bibliographystyle{ieee}
\bibliography{cite}
}

\end{document}